\documentclass[runningheads]{llncs}

\usepackage[utf8]{inputenc}
\usepackage[T1]{fontenc}

\usepackage{graphicx}

\usepackage{amsmath,amssymb} 
\usepackage{color}
\usepackage{wrapfig}

\usepackage{booktabs}       
\usepackage{amsfonts}       
\usepackage{nicefrac}       
\usepackage{microtype}      

\usepackage{algorithm,algcompatible}
\usepackage{multirow}
\usepackage[usenames,dvipsnames]{xcolor}

\usepackage[pagebackref,breaklinks,colorlinks,bookmarks=false]{hyperref}
\usepackage[subtle,mathdisplays=normal,wordspacing=normal,trackingfraction=0.985]{savetrees}
\usepackage{bibunits}
\defaultbibliography{splncs04}

\newcommand{\ROne}[1]{{\color{BlueViolet}\textbf{{#1}}}}
\newcommand{\RTwo}[1]{{\color{PineGreen}\textit{#1}}}

\newcommand{\IGNORE}[1]{}

\newcommand{\PST}[1]{{\color{Green}{#1}}}
\newcommand{\NST}[1]{{\color{RedOrange}{#1}}}

\newcommand{\BET}[1]{{\color{Green}{#1}}}
\newcommand{\WOR}[1]{{\color{RedOrange}{#1}}}
\newcommand{\PAR}[1]{{{#1}}}

\DeclareMathOperator{\diag}{diag}
\DeclareMathOperator{\vecop}{vec}

\usepackage{mathtools}
\usepackage{xspace}
\usepackage{bm}

\newcommand{\mba}{\mathbf{a}}
\newcommand{\mbb}{\mathbf{b}}
\newcommand{\mbc}{\mathbf{c}}
\newcommand{\mbe}{\mathbf{e}}
\newcommand{\mbf}{\mathbf{f}}
\newcommand{\mbg}{\mathbf{g}}
\newcommand{\mbh}{\mathbf{h}}

\newcommand{\mbm}{\mathbf{m}}

\newcommand{\mbw}{\mathbf{w}}
\newcommand{\mbx}{\mathbf{x}}

\newcommand{\R}{\mathbb{R}}

\newcommand{\calE}{\mathcal{E}}
\newcommand{\calF}{\mathcal{F}}
\newcommand{\calG}{\mathcal{G}}
\newcommand{\calH}{\mathcal{H}}

\newcommand{\calK}{\mathcal{K}}
\newcommand{\calL}{\mathcal{L}}
\newcommand{\calM}{\mathcal{M}}

\newcommand{\calO}{\mathcal{O}}
\newcommand{\calP}{\mathcal{P}}

\newcommand{\calX}{\mathcal{X}}
\newcommand{\calY}{\mathcal{Y}}

\def\R{\mathbb{R}}

\def\calX{\mathcal{X}}

\newcommand{\norm}[1]{\left\|#1\right\|}
\newcommand{\abs}[1]{\left|#1\right|}

\def\tr{\mathop{\rm tr}\nolimits}

\def\ones{\mathbf{1}}

\def\eps{\varepsilon}

\newcommand{\ignore}[1]{}

\makeatletter
\DeclareRobustCommand\onedot{\futurelet\@let@token\@onedot}
\def\@onedot{\ifx\@let@token.\else.\null\fi\xspace}
\def\eg{{e.g}\onedot}

\def\ie{{i.e}\onedot}

\def\etal{{et al}\onedot}
\makeatother

\AtBeginDocument{%
\setlength\abovedisplayskip{5pt plus 1pt minus 2.5pt} 
\setlength\belowdisplayskip{5pt plus 1pt minus 2.5pt} 
\setlength\abovedisplayshortskip{0pt plus 3pt} 
\setlength\belowdisplayshortskip{3pt plus 1.5pt minus 1.5pt} 
}

\begin{document}
\pagestyle{headings}
\mainmatter
\def\ECCVSubNumber{100}  

\title{Combining Task Predictors\\
 via Enhancing Joint Predictability} 


\titlerunning{Combining Task Predictors via Enhancing Joint Predictability}
%
\author{Kwang In Kim\inst{1}\orcidID{0000-0002-6470-4571} \and
Christian Richardt\inst{2}\orcidID{0000-0001-6716-9845} \and
Hyung Jin Chang\inst{3}\orcidID{0000-0001-7495-9677}}
\authorrunning{Kim et al.}
%
\institute{\noindent$^1$\enspace UNIST, Korea \quad $^2$\enspace University of Bath, UK \quad $^3$\enspace University of Birmingham, UK}
\maketitle
\begin{abstract}
Predictor combination aims to improve a (target) predictor of a learning task based on the (reference) predictors of potentially relevant tasks, without having access to the internals of individual predictors. 
We present a new predictor combination algorithm that improves the target by i) measuring the relevance of references based on their capabilities in predicting the target, and ii) strengthening such estimated relevance.
Unlike existing predictor combination approaches that only exploit pairwise relationships between the target and each reference, and thereby ignore potentially useful dependence among references, our algorithm \emph{jointly} assesses the relevance of all references by adopting a Bayesian framework.
This also offers a rigorous way to automatically select only relevant references. 
Based on experiments on seven real-world datasets from visual attribute ranking and multi-class classification scenarios, we demonstrate that our algorithm offers a significant performance gain and broadens the application range of existing predictor combination approaches.
\end{abstract}

\section{Introduction}

Many practical visual understanding problems involve learning multiple tasks.
When a \emph{target predictor}, \eg a classification or a ranking function tailored for the task at hand, is not accurate enough, one could benefit from knowledge accumulated in the predictors of other tasks (\emph{references}).
The \textbf{\emph{predictor combination} problem} studied by Kim \etal \cite{KimTomRic17} aims to improve the target predictor by exploiting the references without requiring access to the internals of any predictors or assuming that all predictors belong to the same class of functions.
This is relevant when the forms of predictors are not known (\eg precompiled binaries) or the best predictor forms differ across tasks. For example, Gaussian process rankers~\cite{Joachims2002} trained on ResNet101 features \cite{HeZhaRen16} are identified as the best for the main task, \eg for image frame retrieval, while convolutional neural networks are presented as a reference, \eg classification of objects in images.
In this case, existing transfer learning or multi-task learning approaches, such as a parameter or weight sharing, cannot be applied directly.



Kim~\etal~\cite{KimTomRic17} approached this predictor combination problem
for the first time by nonparametrically accessing all predictors based on their evaluations on given datasets, regarding each predictor as a Gaussian process (GP) estimator.
Assuming that the target predictor is a noisy observation of an underlying ground-truth predictor, their algorithm projects all predictors onto a Riemannian manifold of GPs and denoises the target by simulating a diffusion process therein.
This approach has demonstrated a noticeable performance gain while meeting the challenging requirements of the predictor combination problem.
However, it leaves three possibilities to improve.
Firstly, this algorithm is inherently (pairwise) metric-based and, therefore, it can model and exploit only pairwise relevance of the target and each reference, while relevant information can lie in the relationship between multiple references.
Secondly, this algorithm assumes that all references are noise-free, while in practical applications, the references may also be trained based on limited data points or weak features and thus they can be imperfect.
Thirdly, as this algorithm uses the metric defined between GPs, it can only combine one-dimensional target and references.

In this paper, we propose a new predictor combination algorithm that overcomes these three challenges. The proposed algorithm builds on the manifold denoising framework \cite{KimTomRic17} but instead of their metric diffusion process, we newly pose the predictor denoising as an averaging process, which \emph{jointly} exploits \emph{full dependence} of the references.
%
%
%
%
Our algorithm casts the denoising problem into 1) measuring the \emph{joint} capabilities of the references in predicting the target, and 2) optimizing the target as a variable to enhance such prediction capabilities.
By adopting Bayesian inference under this setting, identifying relevant references is now addressed by a rigorous Bayesian relevance determination approach.
Further, by denoising \emph{all} predictors in a single unified framework, our algorithm becomes applicable even
for imperfect references.
Lastly, our algorithm can combine multi-dimensional target and reference predictors, \eg it can improve multi-class classifiers based on one-dimensional rank predictors.
Experiments on \emph{relative attribute} ranking and multi-class classification demonstrate that these contributions individually and collectively improve the performance of predictor combination and further extend the application domain of existing predictor combination algorithms.

\textit{\textbf{Related work.}}
Transfer learning (TL) aims to solve a given learning problem by adapting a source model trained on a different problem \cite{PanYan10}. Predictor combination can be regarded as a specific instance of TL. However, unlike predictor combination algorithms, traditional TL approaches improve or newly train predictors of \emph{known} form.
Also, most existing algorithms assume that the given source is relevant to the target and, therefore, they do not explicitly consider identifying relevant predictors among many (potentially irrelevant) source predictors.

Another related problem is multi-task learning (MTL), which learns predictors on multiple problems at once~\cite{ArgEvgPon08,CheZhaLi14}. State-of-the-art MTL algorithms offer the capability of automatically identifying relevant task groups when not all tasks and the corresponding predictors are mutually relevant. For example, Argyriou~\etal~\cite{ArgEvgPon08} and Gong~\etal~\cite{GonYeZhang12}, respectively, enforced the sparsity and low-rank constraints in the parameters of predictors to make them aggregate in relevant task groups. Passos~\etal~\cite{PasRaiWai12} performed explicit task clustering ensuring that all tasks (within a cluster) that are fed to the MTL algorithm are relevant. More recently, Zamir \etal \cite{ZamSaxShe18} proposed to discover a hypergraph that reveals the interdependence of multiple tasks and facilitates transfer of knowledge across relevant tasks.

\begin{figure*}[t]
	\centering
	\includegraphics[width=0.88\linewidth]{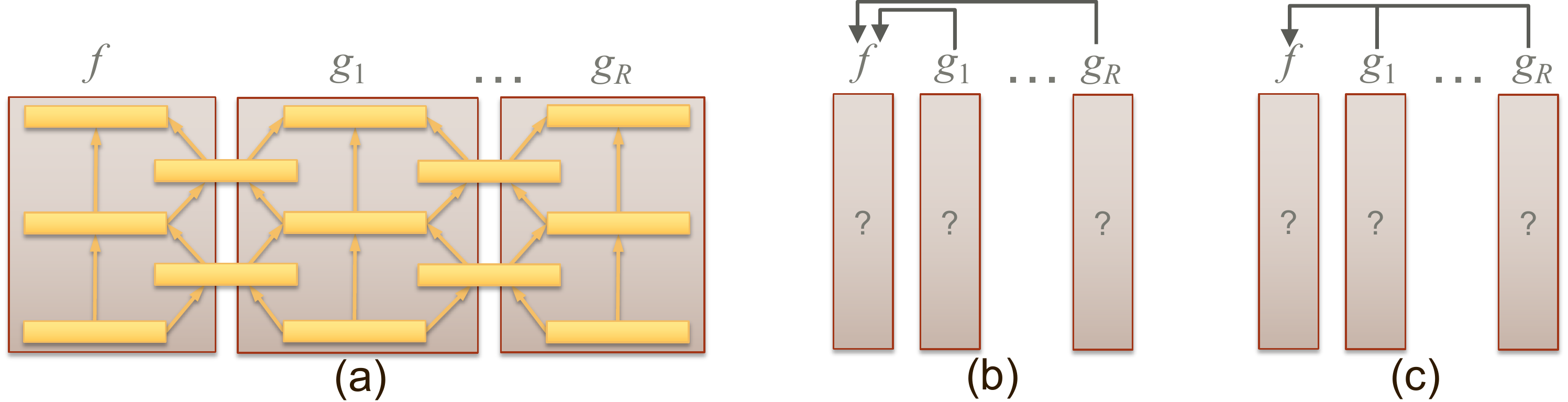}
	\caption{%
		Illustration of predictor combination algorithms:
		(a) MTL simultaneously exploits all references $\{g_1,\ldots,g_R\}$ to improve the target predictor $f$, \eg by sharing neural network layers.
		However, they require access to the internals of predictors~\cite{CheZhaLi14}.
		(b) Kim~\etal's predictor combination is agnostic to the forms of individual predictors~\cite{KimTomRic17} but exploits only pairwise relationships.
		(c) Our algorithm combines the benefits of both, jointly exploiting all references without requiring their known forms.}
	\label{f:diagram}
\end{figure*}

While our approach has been motivated by the success of TL and MTL approaches, these approaches are not directly applicable to predictor combination as they share knowledge across tasks via the internal parametric representations \cite{ArgEvgPon08,GonYeZhang12,PasRaiWai12} and/or shared deep neural network layers of all predictors (\eg via shared encoder readouts \cite{ZamSaxShe18}; see Fig.~\ref{f:diagram}).
%
%
A closely related approach in this context is Mejjati~\etal's nonparametric MTL approach~\cite{MejCosKim18}.
Similar to Kim \etal \cite{KimTomRic17}, this algorithm assesses predictors based on their sample evaluations, and it (nonparametrically) measures and enforces pairwise statistical dependence among predictors.
As this approach is agnostic to the forms of individual predictors, it can be adapted for predictor combination.
However, this algorithm shares the same limitations: 
it can only model pairwise relationships.
We demonstrate experimentally that by modeling the joint relevance of all references, our algorithm can significantly outperform both Kim~\etal's original predictor combination algorithm \cite{KimTomRic17} adapted to ranking~\cite{KimCha19}, and Mejjati~\etal's MTL algorithm~\cite{MejCosKim18}.


\section{The predictor combination problem}
\label{s:opcanalysis}

Suppose we have an \emph{initial predictor} $f^0 \colon \calX\to\calY$ (\eg a classification, regression, or ranking function) of a task. The goal of predictor combination is to improve the \emph{target predictor} $f^0$ based on a set of \emph{reference predictors} $\calG=\{g_i\colon\calX\to\calY_i\}_{i=1}^R$.  
The internal structures of the target and reference predictors are unknown and they might have different forms.
Crucial to the success of addressing this seriously ill-posed problem is to determine which references (if any) within $\calG$ are \emph{relevant} (\ie useful in improving $f^0$), and to design a procedure that fully exploits such relevant references without requiring access to the internals of $f^0$ and $\calG$. 

Kim et al.'s original predictor combination (\emph{OPC}) \cite{KimTomRic17} approaches this problem by 1) considering the initial predictor $f^0$ as a noisy estimate of the underlying ground-truth $f_\text{GT}$, and 2) assuming $f_\text{GT}$ and $\calG$ are structured such that they all lie on a low-dimensional predictor manifold $\calM$.
These assumptions enable predictor combination to be cast as well-established \emph{Manifold Denoising}, where one iteratively denoises points on $\calM$ via simulating the diffusion process therein~\cite{HeiMai07}.


The model space $\calM$ of \emph{OPC}
consists of Bayesian estimates: each predictor in $\calM$ is a GP predictive distribution of the respective task.
The natural metric $g_\calM$ on $\calM$, in this case, is induced from the Kullback-Leibler (KL) divergence $D_\text{KL}$ between probability distributions.
Now further assuming that all reference predictors are noise-free, their diffusion process is formulated as a time-discretized evolution of $f^t$ on $\calM$:
Given the solution $f^t$ at time $t$ and noise-free references $\calG$, the new solution $f^{t+1}$ is obtained by minimizing the energy
\begin{align}
\label{e:ttc}
\calE_\text{O}(f) &= D_\text{KL}^2(f \mid f^t) + \lambda_\text{O}\sum_{i=1}^R w_i D_\text{KL}^2(f \mid g_i) \text{,}
\end{align}
%
where $w_i \!=\! \exp(-D_\text{KL}^2(f^t \mid g_i)/\sigma^2_\text{O})$ is inversely proportional to $D_\text{KL}(f^t \mid g_i)$, and $\lambda_\text{O}$ and $\sigma^2_\text{O}$ 
are hyperparameters. Our supplemental document presents how the iterative minimization of $\calE_\text{O}$ is obtained by discretizing the diffusion process on $\calM$. 

In practice, it is infeasible to directly optimize functions, which are infinite-dimen\-sional objects. Instead, \emph{OPC} approximates all predictors $\{f, \calG\}$ via their evaluations on a test dataset $X \!=\! \{\mbx_1,\ldots,\mbx_N\}$, and optimizes the sample $f$-evaluation $\mbf = f|_X := [f(\mbx_1), \ldots, f(\mbx_N)]^\top$ based on the sample references $\calG = \{\mbg_1, \ldots, \mbg_R\}$ with $\mbg_i = g_i|_X$.


At each time step, the relevance of a reference is automatically determined based on its KL-divergence to the current solution: $g_i$ is considered relevant when $D_\text{KL}(f^t \mid g_i)$ is small.
Then, throughout the iteration, \emph{OPC} robustly denoises $f$ by gradually putting more emphasis on highly relevant references while ignoring outliers.
This constitutes the first predictor combination algorithm that improves the target predictor without requiring any known forms of predictors (as the KL-divergences are calculated purely based on predictor evaluations).
However, Eq.~\ref{e:ttc} also highlights the limitations of this approach: it exploits only pairwise relationships between the target predictor and individual references, ignoring the potentially useful information that lies in the dependence between references. 

\textit{\textbf{Toy problem 1.}}
Consider two references, $\{\mbg_1, \! \mbg_2\} \!\subset\! \R^{100}$, constructed by uniformly randomly sampling from $\{0,1\}$.  
Here, $\{\mbg_1, \mbg_2\}$ are regarded as the means of GP predictive distributions with unit variances.
%
We define the ground-truth target as their difference: $\mbf_\text{GT} = \mbg_1 - \mbg_2$.
By construction, $\mbf_\text{GT}$ is determined by the \emph{relationship} between the references.
Now we construct the initial noisy predictor $\mbf^0$ by adding independent Gaussian noise with standard deviation $1$ to $\mbf_\text{GT}$, achieving the rank accuracy of $0.67$ (see Sec.~\ref{s:experiments} for the definition of the visual attribute ranking problem).
In this case, applying \emph{OPC} minimizes $\calE_\text{O}$ (Eq.~\ref{e:ttc}) but 
shows insignificant performance improvement as no information on $\mbf_\text{GT}$ can be gained by assessing the relevance of the references individually (Table~\ref{t:toyproblems}). 
While this problem has been well-studied in existing MTL and TL approaches, the application of these techniques for predictor combination is not straightforward as they require simultaneous training~\cite{GonYeZhang12,PasRaiWai12} and/or shared predictor forms~\cite{ZamSaxShe18}.
%
%
Another limitation is that \emph{OPC} requires that all predictions are one-dimensional (\ie $\calY_i \subset \R$).
Therefore, it is not capable of, for example, improving the multi-class classification predictor $\mbf^0$ given the references constructed for ranking tasks.

\begin{table}[t]
\caption{%
Accuracies of Kim \etal's original (\emph{OPC})~\cite{KimTomRic17}, and our
linear (\emph{LPC}) and nonlinear (\emph{NPC}) predictor combination algorithms introduced in Section~\ref{s:predictorcombination},
for illustrative toy problems.
$\mbg_1$ and $\mbg_2$ are random binary vectors while $\mbf^0$'s are noisy observations of the corresponding ground-truth predictors $\mbf_\text{GT}$'s. 
}
\label{t:toyproblems}
\setlength{\tabcolsep}{6pt}
\centering
\begin{tabular}{lcccc}
\toprule
Toy problem& $\mbf^0$ & \emph{OPC}~\cite{KimTomRic17} (Eq.~\ref{e:ttc})&    \emph{LPC} (Eq.~\ref{e:lpc}) &     \emph{NPC} (Eq.~\ref{e:objectivef})\\
\midrule
1: $\mbf_\text{GT}=\mbg_1-\mbg_2$&67.14&67.24 
&\textbf{100}&\textbf{100}\\
2: $\mbf_\text{GT}=\text{XOR}(\mbg_1,\mbg_2)$&74.08&74.11&74.24&\textbf{100}\\
\bottomrule
\end{tabular}%
\end{table}

\section{Joint predictor combination algorithm}
\label{s:predictorcombination}

Our algorithm takes deterministic predictors instead of Bayesian predictors (\ie GP predictive distributions) as in \emph{OPC}.
When Bayesian predictors are provided as inputs, we simply take their means and discard the predictive variances.
This design choice offers a wider range of applications as most predictors – including deep neural networks and support vector machines (SVMs) – are presented as deterministic functions, at the expense of not exploiting potentially useful predictive uncertainties.
This assumption has also been adopted by Kim and Chang \cite{KimCha19}.
Under this setting, our model space is a sub-manifold $\calM$ of $L^2$ space where each predictor has zero mean and unit norm:
\begin{align}
\label{e:normalization}
\forall f\in \calM. \quad \int f(\mbx) \mathrm{d}P(\mbx)=0 \quad \text{and} \quad \langle f,f \rangle =1 \text{,}
\end{align}
where $\langle f, g \rangle := \int f(\mbx)g(\mbx) dP(\mbx)$ and $P(\mbx)$ is the probability distribution of $\mbx$.
This normalization enables scale and shift-invariant assessment of the relevance of references. 
The Riemannian metric $g_\calM$ on $\calM$ is defined as the \emph{pullback} metric of the ambient $L^2$ space: when $\calM$ is embedded into $L^2$ via the embedding $\imath$,
$g_\calM(a,b) := \langle\imath(a),\imath(b)\rangle$.
\emph{OPC} (Eq.~\ref{e:ttc}) can be adapted for $\calM$ by iteratively maximizing
the objective $\calO_\text{O}$ that replaces the KL-divergence $D_\text{KL}$ with $g_\calM(\cdot,\cdot)$:
\begin{align}
\calO_\text{O}(f) &= g_\calM(f,f^t)^2+\lambda_\text{O}\sum_{i=1}^R w_i g_\calM(f,g_i)^2\text{.} 
\label{e:ttcadapt}
\end{align}
%
For simplicity of exposition, we here assume that the output space is one-dimensional (\ie $\calY_i=\R$).
In Sec.~\ref{s:experiments}, we show how this framework can be extended to multi-dimensional outputs such as for multi-class classification.

\textit{\textbf{The averaging process on $\cal\calM$.}}
Both \emph{OPC} (Eq.~\ref{e:ttc}) and its adaptation to our model space (Eq.~\ref{e:ttcadapt}) can model only the pairwise relationship between the target $f$ 
and each reference $g_i\in \calG$, while ignoring the dependence present across the references (\emph{joint relevance} of $\calG$ on $f$).
We now present a general framework that can capture such joint relevance by iteratively maximizing the objective 
\begin{align}
\calO_\text{J}(f) &= \langle \imath(f), \imath(f^t) \rangle^2 + \lambda_\text{J} \langle \imath(f), \calK[\imath(f)] \rangle \text{,}
\label{e:baseline}
\end{align}
where $\lambda_\text{J}\geq 0$ is a hyperparameter. The linear, non-negative definite averaging operator $\calK \colon \imath(\calM) \to \imath(\calM)$ is responsible to capture the joint relevance of $\calG$ on $f$. 
Depending on the choice of $\calK$, $\calO_\text{J}$ can accommodate a variety of predictor combination scenarios, including $\calO_\text{O}$ as a special case
for $\calK[\imath(f)]=\sum_{i=1}^R  \imath(g_i) w_i \langle \imath(f), \imath(g_i)\rangle$. 

\subsection{Linear predictor combination (LPC)}
\label{sec:LPC}

Our linear predictability operator $\calK_\text{L}$ is defined as\footnote{Here, the term `linear' signifies the capability of $\calK_\text{L}$ to capture the linear dependence of references, independent of $\calK_\text{L}$ being a linear operator as well. 
}
%
\begin{align}
\label{e:linearpredopr}
\calK_\text{L}[\imath(f)] = \sum_{i,j=1}^R \imath(g_i) C^{-1}_{[i,j]} \langle \imath(g_j), \imath(f)\rangle
\end{align}
%
using the \emph{correlation matrix} 
$C_{[i,j]} = \langle \imath(g_i), \imath(g_j)\rangle$.
Interpreting 
$\calK_\text{L}$ becomes straightforward when substituting $\calK_\text{L}$ into the second term of $\calO_\text{J}$ (Eq.~\ref{e:baseline}):
\begin{align}
\langle \imath(f), \calK_\text{L}[\imath(f)]\rangle = \mbc^\top C^{-1} \mbc \text{,}
\end{align}
%
where $\mbc = [\langle \imath(f),\imath(g_1) \rangle,\ldots,\langle \imath(f),\imath(g_R) \rangle]^\top$.
As each predictor in $\calM$ is centered and normalized, all diagonal elements of the correlation matrix $C$ are $1$. The off-diagonal elements of $C$ then represent the dependence among the references, making 
$\langle \imath(f), \calK_\text{L}[\imath(f)]\rangle$ a measure of \emph{joint correlation} between $f$ and $\calG = \{g_i\}_{i=1}^R$.

In practice, $f$ and $\{g_i\}$ might not be originally presented as embedded elements $\imath(f)$ and $\{\imath(g_i)\}$ of $\calM$: \ie they are not necessarily centered or normalized (Eq.~\ref{e:normalization}).
Also, as in the case of \emph{OPC}, it would be infeasible to manipulate infinite-dimensional functions directly.
Therefore, we also adopt sample approximations $\{\mbf,\mbg_1,\ldots,\mbg_R\}$ and explicitly project them onto $\calM$ via normalization: $\mbf \to \overline{\mbf} := \frac{C_N\mbf}{\|C_N\mbf\|}$, where $C_N \!=\! \ones_{N \times N}/N$,
$\ones_{N \times N}$ is an $N \!\times\! N$ matrix of ones, for the sample size $N \!=\! |X|$.
%
For this scenario, we obtain our linear predictor combination (\emph{LPC}) algorithm by substituting Eq.~\ref{e:linearpredopr} into Eq.~\ref{e:baseline}, and replacing $f$, $f^t$, and $g_j$ by $\overline{\mbf}$, $\overline{\mbf}^t$, and $\overline{\mbg}_j$, respectively:
%
\begin{align}
\calO_\text{L}(\mbf) &= \frac{(\mbf^\top \overline{\mbf^t})^2}{\mbf^\top C_N \mbf}+\lambda_\text{J}\calP_\text{L} \text{,}
\label{e:lpc}
\end{align}
where $\calP_\text{L}=\frac{\mbf^\top Q \mbf}{\mbf^\top C_N \mbf}$, $Q=G(G^\top G)^{-1}G$, and $G=[\overline{\mbg}_1,\ldots,\overline{\mbg}_R]$. Here, we pre-projected $\calG$ and $\mbf^t$ onto $\calM$ while $\mbf$ is explicitly projected in Eq.~\ref{e:lpc}.
Note that our goal is not to simply calculate $\calP_\text{L}$ for a fixed $\mbf$, but to optimize $\mbf$ while enhancing $\calP_\text{L}$.

Exploiting the joint relevance of references, \emph{LPC} can provide significant accuracy improvements over \emph{OPC}. For example, \emph{LPC} can generate perfect predictions in Toy Problem 1 (Table~\ref{t:toyproblems}). However, its capability in measuring the joint relevance is limited to linear relationships only. This can be seen by rewriting $\calP_\text{L}$ explicitly in $\mbf$ and $G$:
\begin{align}
\label{e:linearpred}
\calP_\text{L} = \frac{\mbf^\top Q \mbf}{\mbf^\top C_N \mbf}
= 1 - \frac{\sum_{i=1}^N(\mbf_i - q(G_{[i,:]}))^2}{\sum_{i=1}^N(\mbf_i - \sum_{j=1}^n\mbf_j / N)^2} \text{,}
\end{align}
%
where $G_{[i,:]}$ represents the $i$-th row of $G$, and $q(\mba) = \mbw_q^\top \mba$ is the linear function whose weight vector $\mbw_q = (G^\top G)^{-1}G\mbf$ is obtained from least-squares regression that takes the reference matrix $G$ as training input and the target predictor variable $\mbf$ as corresponding labels. 
Then, $\calP_\text{L}$ represents the normalized prediction accuracy: the normalizer $\mbf^\top C_N \mbf$ is simply the variance of $\mbf$ elements. 
For this reason, we call $\calP_\text{L}$ the (linear) \emph{predictability} of $G$ (and equivalently of $\calG$) on $\mbf$. It takes the maximum value of 1 when the linear prediction (made based on $G$) perfectly agrees with $\mbf$ when normalized, and it attains the minimum value 0 when the prediction is no better than taking the mean value of $\mbf$, in which case the mean squared error becomes the variance. Figure~\ref{f:diagram} illustrates our algorithm in comparison with MTL and \emph{OPC}.


\textit{\textbf{Toy problem 2.}}
Under the setting of Toy problem 1, when the target $\mbf_\text{GT}$ is replaced by a variable that is nonlinearly related to the references, \eg using the logical exclusive OR (XOR) of $\mbg_1$ and $\mbg_2$, \emph{LPC} fails to give any noticeable accuracy improvement compared to the baseline $\mbf^0$. 

\subsection{Nonlinear predictor combination (NPC)}
\label{sec:NPC}

Our final algorithm measures the relevance of $\calG$ on $\mbf$ by predicting $\mbf$ via Gaussian process (GP) estimation. 
We use the standard zero-mean Gaussian prior and an i.i.d. Gaussian likelihood with noise variance $\sigma^2$~\cite{RasWill06}.
The resulting prediction is obtained as a Gaussian distribution with mean $\mbm_\mbf$ and covariance $C_\mbf$:
%
\begin{align}
\label{e:gpr}
\kern-0.3em
\mbm_\mbf \!=\! K(K \!+\! \sigma^2I)^{-1}\mbf\text{,}\enskip\;
C_\mbf \!=\! K \!-\! K(K \!+\! \sigma^2 I)^{-1}K \text{,}
\kern-0.3em
\end{align}
%
where $K \in \R^{N\times N}$ is defined using the covariance function $k \colon \R^R\times \R^R\to \R$:
\begin{align}
\label{e:kernel}
K_{[i,j]} = k(G_{[i,:]},G_{[j,:]}) := \exp\left(-\frac{\|G_{[i,:]}-G_{[j,:]}\|^2}{\sigma_k^2}\right) \text{.}
\end{align}
Now we refine the linear predictability $\calP_\text{L}$ by replacing $q(G_{[i,:]})$ in Eq.~\ref{e:linearpred} with the corresponding predictive mean $[\mbm_\mbf]_i$ (where $[\mba]_i$ is the $i$-th element of vector $\mba$):
\begin{align}
\label{e:nonlinearpred}
\calP_\text{N} = \frac{\mbf^\top Q' \mbf}{\mbf^\top C_N\mbf}
= 1 - \frac{\sum_{i = 1}^N([\mbf]_i - [\mbm_\mbf]_i)^2}{\sum_{i=1}^N([\mbf]_i - \sum_{j=1}^N[\mbf]_j / N)^2} \text{,}
\end{align}
where $Q'$ is a positive definite matrix that replaces $Q$ in Eq.~\ref{e:linearpred}:
\begin{align}
Q' = C_N \left(2  K(K + \sigma^2I)^{-1} - (K + \sigma^2I)^{-1}  K  K(K + \sigma^2I)^{-1}\right) C_N \text{.}
\label{e:qprime}
\end{align}
The matrix $Q'$ becomes $Q$ when the kernel $k(\mba,\mbb)$ is replaced by the standard dot product $k'(\mba,\mbb) = \mba^\top\mbb$.
Note that the noise level $\sigma^2$ should be strictly positive; otherwise,
$\mbf_i = [\mbm_\mbf]_i$ for all $i \in \{1,\ldots,N\}$, and therefore $\calP_\text{N} = 1$ for any $\mbf$.
This means the resulting GP model perfectly overfits to $\mbf$ and all references are considered perfectly relevant regardless of the actual values of $G$ and $\mbf$.


\textit{\textbf{Computational model.}}
Explicitly normalizing $\mbf$ ($\mbf \to \overline{\mbf}$) in the \emph{nonlinear predictability} $\calP_\text{N}$ (Eq.~\ref{e:nonlinearpred}), substituting $Q'$ into $\calP_\text{N}$, and then replacing $\calP_\text{L}$ with $\calP_\text{N}$ in $\calO_\text{L}$ (Eq.~\ref{e:lpc}) yields the following Rayleigh quotient-type objective to maximize:
\begin{align}
\label{e:objectivef}
\calO_\text{N}(\mbf) = \frac{\mbf^\top A\mbf}{\mbf^\top C_N \mbf} \text{,} \quad A= (C_N\mbf^t)(C_N\mbf^t)^\top+\lambda_\text{J} Q' \text{.}
\end{align}
For any non-negative definite matrices $A$ and $C_N$, the maximizer of the Rayleigh quotient $\calO_\text{N}$ is the largest eigenvector (the eigenvector corresponding to the maximum eigenvalue) of the generalized eigenvector problem $A\mbf = \lambda C_N\mbf$.
The computational complexity of solving the generalized eigenvector problem of matrices $\{A,C_N\} \subset \R^{N\times N}$ is $O(N^3)$.
As in our case $N=|X|$, solving this problem is infeasible for large-scale problems.
To obtain a computationally affordable solution, we first note that $A$ incorporates multiplications by the centering matrix $C_N$ and, therefore, all eigenvectors of $A$ are centered, which implies that they are also eigenvectors of $C_N$.
This effectively renders the generalized eigenvector problem into the standard eigenvector problem of matrix $A$.

Secondly, we make sparse approximate GP inference by adopting a low-rank approximation of $K$~\cite{SeeIllLaw03}: 
\begin{align}
\label{e:nystromapprox}
& K \approx K_{GB} K_{BB}^{-1} K_{GB}^\top \text{, } 
 {K_{GB}}_{[i,j]} = k(G_{[i,:]},B_{[j,:]}) \text{, }
{K_{BB}}_{[i,j]} = k(B_{[i,:]},B_{[j,:]}) \text{,} 
\end{align}
where the $i$-th row $B_{[i,:]}$ of $B\in \R^{N'\times R}$ represents the $i$-th \emph{basis vector}.
We construct the basis vector matrix $B$ by linearly sampling $N'$ rows from all rows of $G$.
Now substituting the kernel approximation in Eq.~\ref{e:nystromapprox} into Eq.~\ref{e:qprime} leads to
\begin{align}
\label{e:qprime2}
Q'' &=  C_N K_{GB} \left(\lambda_\text{J}T\right) K_{GB}^\top C_N \text{,\quad with} \\
T &= 2 P -P K_{GB}^\top K_{GB} P \text{\quad and\quad} P = (K_{GB}^\top K_{GB} +\lambda K_{BB})^{-1} \text{.}
\end{align}
Replacing $Q'$ in $A$ with $Q''$, we obtain $A= YY^\top$, where
\begin{align}
\label{e:Ydecomp}
Y = \left[C_N\mbf^t, \sqrt{\lambda_\text{J}}K_{GB}T^\frac{1}{2}\right] \in \R^{N\times (N'+1)}
\end{align}
and $T^\frac{1}{2}(T^\frac{1}{2})^\top =T\in \R^{N'\times N'}$.
Note that $T$ is positive definite (PD) for $\sigma^2>0$ as $Q''$ is PD, which can be seen by noting that $0 \leq \frac{\mbf^\top Q'' \mbf}{\mbf^\top C_N \mbf}\leq 1$: by construction, $\mbf^\top Q'' \mbf$ is the prediction accuracy upper bounded by $\mbf^\top C_N \mbf$.
Therefore, $T^\frac{1}{2}$ can be efficiently calculated based on the Cholesky decomposition of $T$.
In the rare case where Cholesky decomposition cannot be calculated, \eg due to round-off errors, we perform the (computationally more demanding) eigenvalue decomposition $E\Lambda E^\top$ of $T$, replace all eigenvalues in $\Lambda$ that are smaller than a threshold $\eps = 10^{-9}$ by $\eps$, and construct $T^\frac{1}{2}$ as $E\Lambda^\frac{1}{2}$.

Finally, by noting that, when normalized, the largest eigenvector of $YY^\top \in \R^{N\times N}$ is the same as $Y\mbe$, where $\mbe$ is the largest eigenvector of $Y^\top Y\in \R^{(N'+1)\times (N'+1)}$, the optimum $\mbf^*$ of $\calO_\text{N}$ in Eq.~\ref{e:objectivef} is obtained as $\frac{Y\mbe}{\|Y\mbe\|}$ and $\mbe$ can be efficiently calculated by iterating the power method on $Y^\top Y$.
The normalized output $\mbf^*$ can be directly used in some applications, \eg ranking.
When the absolute values of predictors are important, \eg in regression and multi-class classification, the standard deviation and the mean of $\mbf^0$ can be stored before the predictor combination process and $\mbf^*$ is subsequently inverse normalized.

\subsection{Automatic identification of relevant tasks}

Our algorithm \emph{NPC} is designed to exploit all references.
However, in general, not all references are relevant and therefore, the capability of identifying only relevant references can help.
\emph{OPC} does so by defining the weights $\{w_i\}$ (Eq.~\ref{e:ttc}).
However, this strategy inherits the limitation of \emph{OPC} in that it does not consider all references jointly.
An important advantage of our approach, formulating predictor combination as enhancing the predictability via Bayesian inference, is that the well-established methods of automatic relevance determination can be employed for identifying relevant references.
In our GP prediction framework, the contributions of references are controlled by the kernel function $k$ (Eq.~\ref{e:kernel}).
The original Gaussian kernel $k$ uses (isotropic) Euclidean distance $\|\cdot\|$ on $\calX$ and thus treats all references equally.
Now replacing it by an \emph{anisotropic} kernel
\begin{align}
\label{e:anisotropickernel}
k_A(\mba,\mbb)=\exp\left(-(\mba-\mbb)^\top\Sigma_A (\mba-\mbb)\right)    
\end{align}
%
with $\Sigma_A = \text{diag}\left[\sigma_A^1,\ldots,\sigma_A^R\right]$ being a diagonal matrix of non-negative entries renders the problem of identifying relevant references into estimating the hyperparameter matrix $\Sigma_A$:
when $\sigma_A^i$ is large, then $\mbg_i$ is considered relevant and it makes a significant contribution in predicting $\mbf$, while a small $\sigma_A^i$ indicates that $\mbg_i$ makes a minor contribution.

For a fixed target predictor $\mbf$, identifying the optimal parameter $\Sigma_A^*$ is a well-studied problem in Bayesian inference: $\Sigma_A^*$ can be determined by maximizing the \emph{marginal likelihood}~\cite{RasWill06} $p(\mbf \mid G,\Sigma_A)$.
This strategy cannot be directly applied to our algorithm as $\mbf$ is the variable that is optimized depending on the prediction made by GPs.
Instead, one could estimate $\Sigma_A^*$ based on the initial prediction $\mbf^0$ and $G$, and fix it throughout the optimization of $\mbf$.
We observed in our preliminary experiments that this strategy indeed led to noticeable performance improvement over using the isotropic kernel $k$.
However, optimizing the GP marginal likelihood $P(\mbf \mid G,\Sigma_A)$ for a (nonlinear) Gaussian kernel (Eq.~\ref{e:kernel}) is computationally demanding: this process takes roughly 1,000 times longer than the optimization of $\calO_\text{N}$ (Eq.~\ref{e:objectivef}; for the \emph{AWA2} dataset case; see~Sec.~\ref{s:experiments}).
Instead, we first efficiently determine surrogate parameters $\Sigma_L = \text{diag}\left[\sigma_L^1,\ldots,\sigma_L^R\right]$ by optimizing the marginal likelihood based on the linear anisotropic kernel $k_L(\mba, \mbb) = \mba^\top \Sigma_L \mbb$.
In our preliminary experiments, we observed that once optimized, the relative magnitudes of $\Sigma_L^*$ elements are similar to these of $\Sigma_A^*$, but their global scales differ (see the supplemental document
for examples and details of marginal likelihood optimization).
In our final algorithm, we determine $\Sigma_A^*$ by 
scaling $\Sigma_L^*$: $\Sigma_A^* = \Sigma_L^* / \sigma_k^2$ for a hyperparameter $\sigma_k^2 > 0$.

\begin{figure}[t]
\centering
\includegraphics[width=0.55\columnwidth]{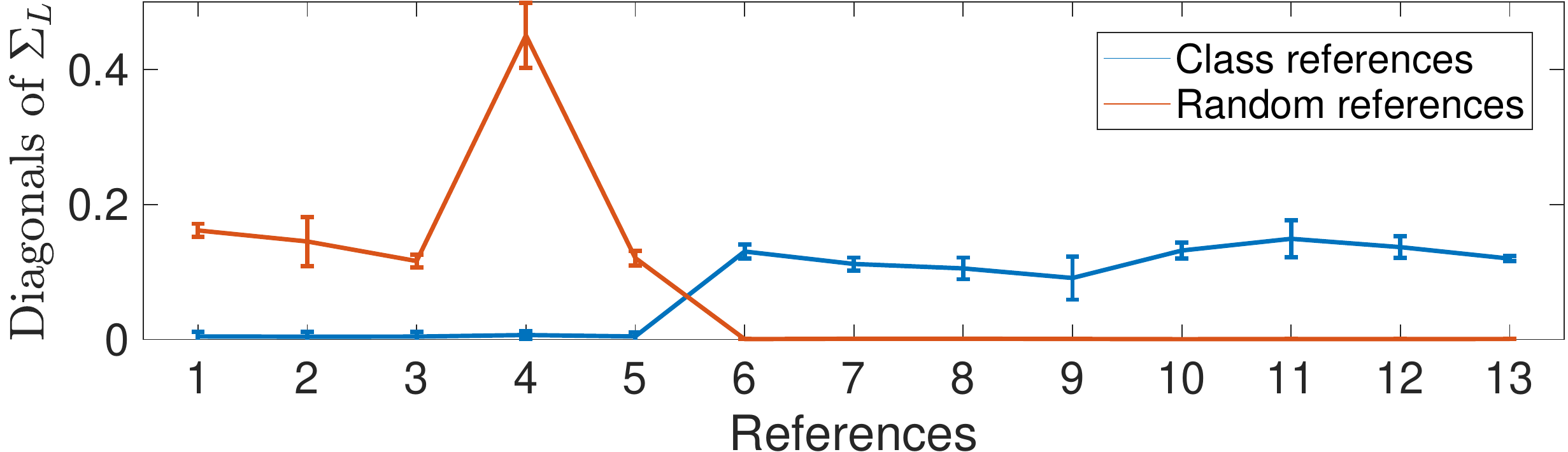}
\caption{%
	The average diagonal values of $\Sigma_L^*$ optimized for the first attribute of the \emph{OSR} dataset as the target with remaining 5 attributes in the same dataset as references 1 to 5, plus 8 additional attributes as references 6 to 13.
	$\Sigma_L^*$ values are normalized to sum to one for visualization.
	The length of each error bar corresponds to twice the standard deviation.
	\emph{Class references}: References 6–13 are class labels from which attribute labels are generated.
	\emph{Random references}: References 6–13 are randomly generated.
	See text.
	}
\label{f:sigmaml}
\end{figure}

Figure~\ref{f:sigmaml} demonstrates the effectiveness of automatic relevance determination: The \emph{OSR} dataset contains 6 target attributes for each data instance, which are defined based on the underlying class labels.
The figure shows the average diagonal values of $\Sigma_L^*$ on this dataset estimated for the first attribute using the remaining 5 attributes, plus 8 additional attributes as references.
Two scenarios are considered.
In the \emph{random references} scenario, the additional attributes are randomly generated. As indicated by small magnitudes and the corresponding standard deviations of $\Sigma_L^*$ entries, our algorithm successfully disregarded these irrelevant references.
In \emph{class references} scenario, the additional attributes are ground-truth class labels which provide complete information about the target attributes.
Our algorithm successfully \emph{picks up} these important references.
On average, removing the automatic relevance determination from our algorithm decreases the accuracy improvement (from the initial predictors $\mbf^0$) by 11.97\% (see Table~\ref{t:ablation}).


\subsection{Joint denoising}

So far, we assumed that all references in $\calG$ are noise-free.
However, in practice, they might be noisy estimates of the ground truth.
In this case, noise in the references could be propagated to the target predictor during denoising, which would degrade the final output.
We account for this by denoising \emph{all} predictors $\calH = \{\mbf, \mbg_1,\ldots,\mbg_R\}$ simultaneously.
At each iteration $t$, each predictor $\mbh \in \calH$ is denoised by considering it as the target predictor, and $\calH \setminus \{\mbh\}$ as the references in Eq.~\ref{e:objectivef}.
In the experiments with the \emph{OSR} dataset, removing this joint denoising process from our final algorithm decreases the average accuracy rate by 8.26\% (see Table~\ref{t:ablation}).
We provide a summary of our complete algorithm in the supplemental document.


\begin{table}[t]
\caption{%
	Effect of design choices in our algorithm on the \emph{OSR} dataset.
	The average rank accuracy improvement over multiple target attributes from the baseline initial predictions $\mbf^0$ are shown (see Sec.~\ref{s:experiments} for details).
	\emph{w/o joint denois.} only denoising the target predictor.
	\emph{w/o auto. relev.}: without automatic relevance determination.
	Numbers in parentheses are accuracy ratios w.r.t. \emph{Final NPC}.}
\label{t:ablation}
\setlength{\tabcolsep}{6pt}
\centering
\renewcommand*{\arraystretch}{1.1}%
\begin{tabular}{rccc}
	\toprule
	Design choices $\to$ & \emph{w/o joint denois.} & \emph{w/o auto. relev.} & \emph{Final NPC} \\ \midrule
	Accuracy improvement &      1.96 (91.74\%)      &     1.88 (88.03\%)      & \textbf{2.13 (100\%)} \\ \bottomrule
\end{tabular}
\end{table}

\textit{\textbf{Computational complexity and discussion.}}
Assuming that $N\gg R$, the computational complexity of our algorithm (Eq.~\ref{e:objectivef}) is dominated by calculating the kernel matrix $K_{GB}$ (Eq.~\ref{e:nystromapprox}),
which takes $O(NN'R)$ for $N$ data points, $N'$ basis vectors and $R$ references.
The second-most demanding part is the calculation of $T^\frac{1}{2}$ from $T$ based on Cholesky decomposition (Eq.~\ref{e:qprime2}; $O(N'^3)$).
As we denoise not only the target predictor but also all references, the overall computational complexity of each denoising step is $O(R \times (NN'R + N'^3))$.
%
On a machine with an Intel Core i7 9700K CPU 
and an NVIDIA GeForce RTX 2080 Ti GPU, the entire denoising process, including optimization of $\{(\Sigma_A)_i\}_{i=1}^R$ (Eq.~\ref{e:anisotropickernel}), took around 10 seconds for the \emph{AWA2} dataset with 37,322 data points and 79 references for each target attribute.
For simplicity, we use the low-rank approximation of $K$ (Eq.~\ref{e:nystromapprox}) for constructing sparse GP predictions, while more advanced methods exist~\cite{RasWill06}.
The number $N'$ of basis vectors is fixed at 300 throughout our experiments.
While the accuracy of low-rank approximation (Eq.~\ref{e:nystromapprox}) is in general positively correlated with $N'$, we have not observed any significant performance gain by raising $N'$ to 1,000 in 
our experiments.
GP predictions also generally improve when \emph{optimizing} the basis matrix $B$, \eg via the marginal likelihood~\cite{SnelGhah06} instead of being selected from datasets as we did.
Our efficient eigenvector calculation approach (Eq.~\ref{e:Ydecomp}) can still be applied in these cases.


\section{Experiments}
\label{s:experiments}

We assessed the effectiveness of our predictor combination algorithm in two scenarios:
1) visual attribute ranking \cite{ParGra11}, and
2) multi-class classification guided by the estimated visual attribute ranks.
%
Given a database of images $X \subset \calX$, visual attribute ranking aims to introduce a linear ordering of entries in $X$ based on the strength of semantic attributes present in each image $\mbx \in X$.
For a visual attribute, our goal is to estimate a rank predictor $f\colon \calX \to \R$, such that $f(\mbx_i) > f(\mbx_j)$ when the attribute is stronger in $\mbx_i$ than $\mbx_j$.
Parikh and Grauman's original relative attributes algorithm \cite{ParGra11} estimates a linear rank predictor $f(\mbx) = \mbw^\top \mbx$ via rank SVMs that use the rank loss $\calL$  
defined on ground-truth ranked pairs $U\subset X\times X$:
\begin{align}
\calE(f) &= \sum_{(\mbx_i, \mbx_j) \in U} \calL(f, (\mbx_i, \mbx_j)) + C \|\mbw\|^2 \text{,} \\
\calL(f, (\mba, \mbb)) &= \max\left(1 - (f(\mba) - f(\mbb)), 0\right)^2 \text{.}
\label{e:rankloss}
\end{align}
Yang~\etal~\cite{YanZhaXu16} and Meng~\etal~\cite{MenAdlKim18} extended this initial work using deep neural networks (\emph{neural rankers}).
Kim and Chang~\cite{KimCha19} extended the original predictor combination framework of Kim~\etal~\cite{KimTomRic17} to rank predictor combination.

\textit{\textbf{Experimental settings.}}
For visual attribute ranking, we use seven datasets, each with annotations for multiple attributes per image.
For each attribute, we construct an initial predictor and denoise it via predictor combination using the predictors constructed for the remaining attributes as the reference.
The initial predictors are constructed by first training 1) neural rankers, 2) linear and 3) non-linear rank SVMs, and 4) semi-supervised rankers that use the iterated graph Laplacian-based regularizer~\cite{ZhoBelSre11}, all using the rank loss $\calL$ (Eq.~\ref{e:rankloss}).
For each attribute, we select the ranker with the highest validation accuracy as \emph{baseline} $\mbf^0 = f|_X$.

We compare our proposed algorithm to:
1) the baseline predictor $\mbf^0$,
2) Kim and Chang's adaptation~\cite{KimCha19} of Kim \etal's predictor combination approach \cite{KimTomRic17} to visual attribute ranking (\emph{OPC}), and
3) Mejjati~\etal's multi-task learning (\emph{MTL}) algorithm~\cite{MejCosKim18}.
While the latter was originally designed for MTL problems, it does not require known forms of individual predictors and can be thus adapted for predictor combination.
In the supplemental document, we also compare with an adaptation of Evgeniou~\etal's graph Laplacian-based MTL algorithm~\cite{EvgMicPon05} to the predictor combination setting, which demonstrates that all predictor combination algorithms outperform na\"{i}ve adaptations of traditional MTL algorithms.

Adopting the experimental settings of Kim \etal \cite{KimCha19,KimTomRic17}, we tune the hyperparameters of all algorithms on evenly-split training and validation sets.
%
Our algorithm requires tuning the noise level $\sigma^2$ (Eq.~\ref{e:qprime}), global kernel scaling $\sigma_k^2$, and the regularization parameter $\lambda_\text{J}$ (Eq.~\ref{e:objectivef}), which are tuned based on validation accuracy.
For the number of iterations $S$, we use 20 iterations and select the iteration number that achieves the highest validation accuracy.
%
The hyperparameters for other algorithms are tuned similarly (see the supplemental material for details).
For each dataset, we repeated experiments 10 times with different training, validation, and test set splits and report the average accuracies. 

The \emph{OSR}~\cite{ParGra11}, \emph{Pubfig}~\cite{ParGra11}, and \emph{Shoes}~\cite{KovParGra12} datasets provide 2688, 772 and 14,658 images each and include rank annotations (\ie strengths of attributes present in images) for 6, 11 and 10 visual attributes, respectively.
The attribute annotations in these datasets were obtained from the underlying class labels.
For example, each image in \emph{OSR} is also provided with a ground-truth class label out of 8 classes.
The attribute ranking is assigned per class-wise comparisons such that all images in a class have stronger (or the same) presence of an attribute than another class.
This implies that the class label assigned for each image completely determines its attributes, while attributes themselves might not provide sufficient information to determine classes.
Similarly, the attribute annotations for \emph{Pubfig} and \emph{Shoes} are generated from class labels out of 8 and 10 classes, respectively.
The input images in \emph{OSR} and \emph{Shoes} are represented as combinations of GIST \cite{OlivaT2001} and color histogram features, while \emph{Pubfig} uses GIST features as provided by the authors \cite{KovParGra12,ParGra11}.
In addition, for \emph{OSR}, we extracted 2,048-dimensional features using ResNet101 pre-trained on ImageNet~\cite{HeZhaRen16} to fairly assess the predictor combination performance when the accuracies of the initial predictors are higher thanks to advanced features (\emph{OSR (ResNet)}).

The \emph{aPascal} dataset is constructed based on the \textsc{Pascal} VOC 2008 dataset~\cite{EveEslGoo15} containing 12,695 images with 64 attributes~\cite{FarEndHoi09}. 
Each image is represented as a 9,751-dimensional feature vector combining histograms of local texture, HOG, and edge and color descriptors.
The Caltech-UCSD Birds-200-2011 (\emph{CUB}) dataset \cite{WaBraWelPerBel11} provides 11,788 images with 312 attributes where the images are represented by 
the ResNet101 features.
The Animals With Attributes 2 (\emph{AWA2}) dataset consists of 37,322 images with 85 attributes~\cite{XiaLamSch19}.
We used the ResNet101 features as shared by Xian \etal \cite{XiaLamSch19}.
For \emph{aPascal}, \emph{CUB}, and \emph{AWA2}, the distributions of attribute values are imbalanced.
To ensure that sufficient numbers (300) of training and testing labels exist for each attribute level, we selected 29, 40 and 80 attributes from \emph{aPascal}, \emph{CUB} and \emph{AWA2}, respectively.
The ranking accuracy is measured in 100$\times$ Kendall's rank correlation coefficient, which is defined as the difference between the numbers of correctly and incorrectly ordered rank pairs, respectively, normalized by the number of total pairs (bounded in $100 \times [-1, 1]$; higher is the better).

The UT Zappos50K (\emph{Zap50K}) contains 50,025 images of shoes with 4 attributes. Each image is represented as a combination of GIST and color histogram features provided by Yu and Grauman~\cite{YuGra14}. The ground-truth attribute labels are collected by instance-level pairwise comparison collected via \emph{Mechanical Turk}~\cite{YuGra14}. 


We also performed multi-class classification experiments on the \emph{OSR}, \emph{Pubfig},  \emph{Shoes}, \emph{aPascal}, and \emph{CUB} datasets based on their respective class labels.
The initial predictors $\mbf^0 \colon \calX\to\R^H$ are obtained as deep neural networks with continuous softmax decisions trained and validated on 20 labels per class.
Each prediction is given as an $H$-dimensional vector with $H$ being the number of classes. Our goal is to improve $\mbf^0$ using the predictors for visual attribute ranking as references.
It should be noted that our algorithm \emph{jointly improves} all $H$ class-wise predictors as well as ranking references: 1) all class predictors evolve simultaneously, and 2) for improving the predictor of a class, the (evolving) predictors of the remaining classes are used as additional references.
For a fair comparison, we denoise class-wise predictors using both the rank predictors and the predictors of the remaining classes as references, also for the other predictor combination algorithms. 
\begin{figure*}[t]
\centering
\includegraphics[width=0.95\linewidth]{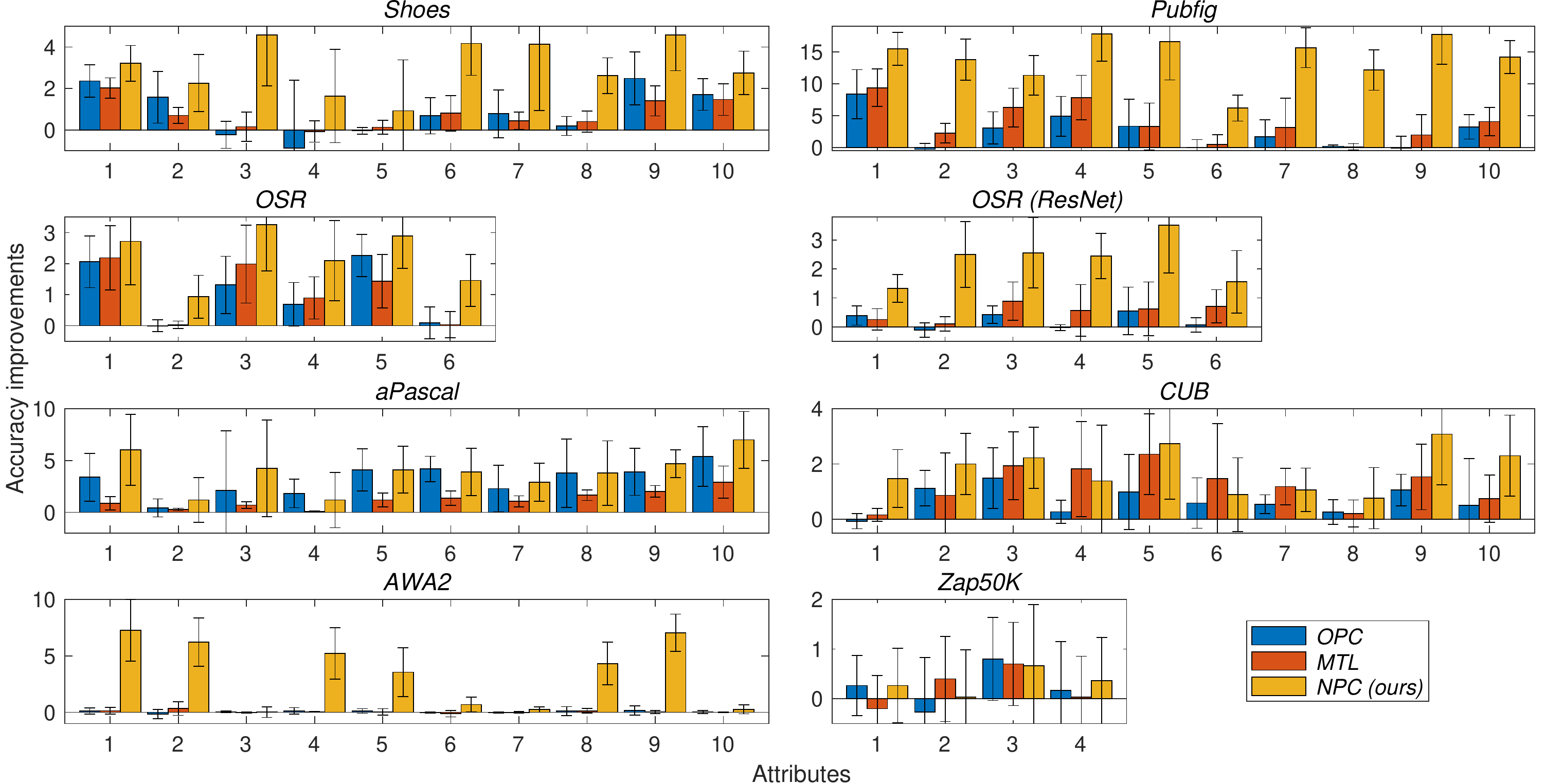}
\caption{%
Average accuracy improvement of different predictor combination algorithms from the \emph{baseline predictors} for up to first 10 attributes. The complete results including statistical significance tests can be found in the supplemental document.}
\label{f:rankingresults}
\end{figure*}
\begin{table}[t]
\caption{%
	Average classification accuracies (\%) using rank estimates as references.
	The numbers in parentheses show the relative accuracy improvement over the baseline $\mbf^0$.
}
\label{t:classresults}
\setlength{\tabcolsep}{6pt}
\centering
 \renewcommand*{\arraystretch}{1.1}%
\resizebox{\columnwidth}{!}{%
%
\begin{tabular}{lccccc}
	\toprule
	                              &     \emph{Shoes}      &     \emph{Pubfig}     &      \emph{OSR}       &      \emph{aPascal} &      \emph{CUB}   \\ \midrule
	Baseline $\mbf^0$             &     57.90 (0.00)      &     77.99 (0.00)      &     76.88 (0.00)      &     37.86 (0.00)    &     66.98 (0.00)  \\ 
	\emph{OPC} \cite{KimCha19}    &     58.52 (1.07)      &     82.55 (5.85)      &     77.16 (0.38)      &     39.77 (5.04)    &     67.75 (1.15)  \\ 
	\emph{MTL} \cite{MejCosKim18} &     59.51 (2.78)      &     80.16 (2.78)      &     77.40 (0.68)      &     38.38 (1.37)    &     67.95 (1.45)  \\ 
	\emph{NPC (ours)}             & \textbf{62.87} (8.58) & \textbf{86.51} (10.9) & \textbf{79.71} (3.69) &     \textbf{40.34} (6.55)    &     \textbf{68.26} (1.92)  \\ \bottomrule
\end{tabular}
}
\end{table}
\textit{\textbf{Ranking results.}}
Figure~\ref{f:rankingresults} summarizes the results for the relative attributes ranking experiments.
Here, we show the results of only the first 10 attributes; the supplemental document contains complete results, which show a similar tendency as presented here.
All three predictor combination algorithms frequently achieved significant performance gains over the baseline predictors $\mbf^0$.
Importantly, apart from one case (
\emph{Shoes} attribute 4
), all predictor combination algorithms did not significantly degrade the performance from the baseline.
This demonstrates the utility of predictor combination.
However, both \emph{OPC} and \emph{MTL} are limited in that they can only capture pairwise dependence between the target predictor and each reference.
By taking into account the dependence present among the references, and thereby \emph{jointly} exploiting them in improving the target predictor, our algorithm further significantly improves the performance:  
Our algorithm performs best for 87.1\% of attributes. 
%
In particular, \emph{Ours} showed significant improvement on 6 out of 10 \emph{AWA2} attributes, where the other algorithms achieved no noticeable performance gain.
This supports our assumption that multiple attributes indeed can \emph{jointly} supply relevant information for improving target predictors, even if not individually.

\textit{\textbf{Multi-class classification results.}}
Table~\ref{t:classresults} shows the results of improving multi-class classifications.
Jointly capturing all rank predictors as well as the multi-dimensional classification predictions as references, our algorithm demonstrates significant performance gains (especially on \emph{Shoes} and \emph{Pubfig}), while other predictor combination algorithms achieved only marginal improvements, confirming the effectiveness of our joint prediction strategy.

\section{Discussions and Conclusions}
\label{s:conclusions}
Our algorithm builds upon the assumption that the reference predictors can help improve the target predictor when they can well predict (or explain) the ground-truth $\mbf_\text{GT}$.
Since $\mbf_\text{GT}$ is not available during testing, we use the noisy target predictor $\mbf^t$ at each time step $t$ as a surrogate, which by itself is iteratively denoised.
While our experiments demonstrate the effectiveness of this approach in real-world examples, simple failure cases exist.
For example, if $\mbf^0$ (as the initial surrogate to $\mbf_\text{GT}$) is contained in the reference set $\calG$, our automatic reference determination approach will pick this up as the single most relevant reference, and therefore, the resulting predictor combination process will simply output $\mbf^0$ as the final result.
We further empirically observed 
that even when the automatic relevance determination is disabled (\ie $\Sigma_L=I$), the performance degraded significantly when $\mbf^0$ is included in $\calG$.
Also, as shown for the \emph{Zap50K} results, there might be cases where no algorithm shows any significant improvement (indicated by the relatively large error bars).
In general, our algorithm may fail when the references do not communicate sufficient \emph{information} for improving the target predictor.
Quantifying such utility of references and predicting the failure cases may require a new theoretical analysis framework.


Existing predictor combination algorithms only consider pairwise relationships between the target predictor and each reference.
This misses potentially relevant information present in the dependence among the references.
We explicitly address this limitation by introducing a new \emph{predictability criterion} that measures how references are \emph{jointly} contributing in predicting the target predictor.
Adopting a fully Bayesian framework, our algorithm can automatically select informative 
references among many potentially irrelevant predictors.
Experiments on seven datasets demonstrated the effectiveness of the proposed predictor combination algorithm.

\noindent\textit{\textbf{Acknowledgements.}}
This work was supported by UNIST's 2020 Research Fund (1.200033.01), National Research Foundation of Korea (NRF) grant NRF-2019\-R1F1A1061603, and Institute of Information \& Communications Technology Planning \& Evaluation (IITP) grant (No.20200013360011001, Artificial Intelligence Graduate School support (UNIST)) funded by the Korean government (MSIT).

\bibliographystyle{splncs04}
\bibliography{./bib/biblio}

\begin{thebibliography}{10}
\providecommand{\url}[1]{\texttt{#1}}
\providecommand{\urlprefix}{URL }
\providecommand{\doi}[1]{https://doi.org/#1}

\bibitem{ArgEvgPon08}
Argyriou, A., Evgeniou, T., Pontil, M.: Convex multi-task feature learning.
  Machine Learning  \textbf{73}(3) (2008)

\bibitem{BoyParChu10}
Boyd, S., Parikh, N., Chu, E., Peleato, B., Eckstein, J.: Distributed
  optimization and statistical learning via the alternating direction method of
  multipliers. Foundations and Trends in Machine Learning  \textbf{3}(1),
  1--122 (2010). \doi{10.1561/2200000016}

\bibitem{CheZhaLi14}
Chen, L., Zhang, Q., Li, B.: Predicting multiple attributes via relative
  multi-task learning. In: CVPR. pp. 1027--1034 (2014)

\bibitem{EveEslGoo15}
Everingham, M., Eslami, S.M.A., Gool, L.V., Williams, C.K.I., Winn, J.,
  Zisserman, A.: The \textsc{Pascal} visual object classes challenge: a
  retrospective. IJCV  \textbf{111}(1),  98--136 (2015)

\bibitem{EvgMicPon05}
Evgeniou, T., Micchelli, C.A., Pontil, M.: Learning multiple tasks with kernel
  methods. JMLR  \textbf{6},  615--637 (2005)

\bibitem{FarEndHoi09}
Farhadi, A., Endres, I., Hoiem, D., Forsyth, D.: Describing objects by their
  attributes. In: CVPR. pp. 1778--1785 (2009)

\bibitem{GonYeZhang12}
Gong, P., Ye, J., Zhang, C.: Robust multi-task feature learning. In: KDD. pp.
  895--903 (2012)

\bibitem{HeZhaRen16}
He, K., Zhang, X., Ren, S., Sun, J.: Deep residual learning for image
  recognition. In: CVPR. pp. 770--778 (2016)

\bibitem{HeiMai07}
Hein, M., Maier, M.: Manifold denoising. In: NIPS. pp. 561--568 (2007)

\bibitem{JitSzaGre15}
Jitkrittum, W., Szab{\'o}, Z., Gretton, A.: An adaptive test of independence
  with analytic kernel embeddings. In: PMLR (Proc. ICML). pp. 1742--1751 (2017)

\bibitem{Joachims2002}
Joachims, T.: Optimizing search engines using clickthrough data. In: KDD. pp.
  133--142 (2002)

\bibitem{KimCha19}
Kim, K.I., Chang, H.J.: Joint manifold diffusion for combining predictions on
  decoupled observations. In: CVPR. pp. 7549--7557 (2019)

\bibitem{KimTomRic17}
Kim, K.I., Tompkin, J., Richardt, C.: Predictor combination at test time. In:
  ICCV. pp. 3553--3561 (2017)

\bibitem{KovParGra12}
Kovashka, A., Parikh, D., Grauman, K.: Whittlesearch: Image search with
  relative attribute feedback. In: CVPR. pp. 2973--2980 (2012)

\bibitem{MejCosKim18}
Mejjati, Y.A., Cosker, D., Kim, K.I.: Multi-task learning by maximizing
  statistical dependence. In: CVPR. pp. 3465--3473 (2018)

\bibitem{MenAdlKim18}
Meng, Z., Adluru, N., Kim, H.J., Fung, G., Singh, V.: Efficient relative
  attribute learning using graph neural networks. In: ECCV. pp. 552--567 (2018)

\bibitem{OlivaT2001}
Oliva, A., Torralba, A.: Modeling the shape of the scene: A holistic
  representation of the spatial envelope. IJCV  \textbf{42}(3),  145--175
  (2001)

\bibitem{PanYan10}
Pan, S.J., Yang, Q.: A survey on transfer learning. IEEE Transactions on
  Knowledge and Data Engineering  \textbf{22}(10),  1345--1359 (2010)

\bibitem{ParGra11}
Parikh, D., Grauman, K.: Relative attributes. In: ICCV. pp. 503--510 (2011)

\bibitem{PasRaiWai12}
Passos, A., Rai, P., Wainer, J., {Daum\'{e} III}, H.: Flexible modeling of
  latent task structures in multitask learning. In: ICML. pp. 1103--1110 (2012)

\bibitem{RasWill06}
Rasmussen, C.E., Williams, C.K.I.: Gaussian Processes for Machine Learning. MIT
  Press, Cambridge, MA (2006)

\bibitem{Sch02}
Sch{\"o}lkopf, B., Smola, A.J.: Learning with Kernels. MIT Press, Cambridge, MA
  (2002)

\bibitem{SeeIllLaw03}
Seeger, M., Williams, C.K.I., Lawrence, N.D.: Fast forward selection to speed
  up sparse {Gaussian} process regression. In: International Workshop on
  Artificial Intelligence and Statistics (2003)

\bibitem{SheMor50}
Sherman, J., Morrison, W.J.: Adjustment of an inverse matrix corresponding to a
  change in one element of a given matrix. The Annals of Mathematical
  Statistics  \textbf{21}(1),  124--127 (1950)

\bibitem{SnelGhah06}
Snelson, E., Ghahramani, Z.: Sparse {Gaussian} processes using pseudo-inputs.
  In: NIPS (2006)

\bibitem{WaBraWelPerBel11}
Wah, C., Branson, S., Welinder, P., Perona, P., Belongie, S.: The {Caltech-UCSD
  Birds-200-2011 Dataset}. Tech. Rep. CNS-TR-2011-001, California Institute of
  Technology (2011)

\bibitem{XiaLamSch19}
Xian, Y., Lampert, C.H., Schiele, B., Akata, Z.: Zero-shot learning -- {A}
  comprehensive evaluation of the good, the bad and the ugly. IEEE TPAMI
  \textbf{41}(9),  2251--2265 (2019)

\bibitem{YanZhaXu16}
Yang, X., Zhang, T., Xu, C., Yan, S., Hossain, M.S., Ghoneim, A.: Deep relative
  attributes. IEEE T-MM  \textbf{18}(9),  1832--1842 (2016)

\bibitem{YuGra14}
Yu, A., Grauman, K.: Fine-grained visual comparisons with local learning. In:
  CVPR. pp. 192--199 (2014)

\bibitem{ZamSaxShe18}
Zamir, A.R., Sax, A., Shen, W., Guibas, L., Malik, J., Savarese, S.: Taskonomy:
  disentangling task transfer learning. In: CVPR. pp. 3712--3722 (2018)

\bibitem{ZhoBelSre11}
Zhou, X., Belkin, M., Srebro, N.: An iterated graph {Laplacian} approach for
  ranking on manifolds. In: KDD. pp. 877--885 (2011)

\end{thebibliography}


\clearpage
\section*{\centering \LARGE Supplemental Document for \\Combining Task Predictors\\
 via Enhancing Joint Predictability}

\noindent
In this supplemental document, we present:
%
\begin{enumerate}
	\item Details of the marginal likelihood calculation used in the automatic determination of relevant predictors $\Sigma_L$ (Sec.~\ref{e:ml});
	\item A summary of our predictor combination algorithm (Sec.~\ref{e:algorithmsummary});
	\item A detailed discussion of baseline algorithms, including:
	\begin{enumerate}
		\item our adaptation of Mejjati~\etal's multi-task learning (\emph{MTL}) algorithm \cite{MejCosKim18} (Sec.~\ref{s:adaptmtl}),
		\item a derivation of Kim~\etal's original predictor combination (\emph{OPC}) algorithm~\cite{KimTomRic17,KimCha19} (Sec.~\ref{s:adaptkim}), and
		\item Evgeniou~\etal's graph Laplacian (\emph{GL})-based MTL algorithm and its adaptation to predictor combination (Sec.~\ref{s:adaptmtl2}).
	\end{enumerate}
\end{enumerate}
In the main paper, we only presented the ranking results for the first 10 attributes in each dataset.
In Section~\ref{s:experiment}, we provide the complete experimental results, including additional results of \emph{GL} and tests of statistical significance of the accuracy improvements made by different algorithms.
We reproduce some content from the main paper to make this document self-contained.

\section{Details of the main algorithm}

\subsection{Calculating the marginal likelihood for linear Gaussian process prediction}
\label{e:ml}

Suppose that we have the following linear and nonlinear anisotropic covariance functions:
\begin{align}
\label{e:linearkernel}
	k_L(\mba, \mbb) &= \mba^\top \Sigma_L \mbb \text{,} \\
	k_A(\mba, \mbb) &= \exp\left(-(\mba - \mbb)^\top \Sigma_A (\mba - \mbb) \right) \text{,}
\label{e:nonlinearkernel}
\end{align}
where $\Sigma_L = \diag[\bm{\sigma}]$, the diagonal matrix with elements $\bm{\sigma} = [\sigma^1, \ldots, \sigma^n]^\top$. $\Sigma_A$ is defined similarly. Our goal is to maximize the marginal likelihood $p(\mbf \mid G, \Sigma_L)$ of the sampled predictor $\mbf$ given the reference matrix $G$ with respect to $\bm{\sigma}$.
The log marginal likelihood $\log(p(\mbf \mid G, \Sigma_L))$ of linear Bayesian regression with Gaussian prior and i.i.d. Gaussian noise model is given as~\cite{RasWill06}:
\begin{align}
	\log(p(\mbf \mid G)) =& -\frac{1}{2} \log \abs{G \cdot \diag[\bm{\sigma}] \cdot G^\top + \lambda I} - \frac{N}{2}\log(2\pi) \nonumber \\
	& -\frac{1}{2}\mbf^\top (G \cdot \diag[\bm{\sigma}] \cdot G^\top + \lambda I)^{-1} \mbf \text{.}
\end{align}
As maximizing $p(\mbf \mid G, \Sigma_L)$ is equivalent to minimizing $-\log(p(\mbf \mid G, \Sigma_L))$, and the second term in $\log(p(\mbf \mid G, \Sigma_L))$ is independent of $\bm{\sigma}$, we can find the optimal parameter vector $\bm{\sigma}^*$ by minimizing the following energy:
\begin{align}
	\calE(\bm{\sigma}) =& \log \abs{G \cdot \diag[\bm{\sigma}] \cdot G^\top + \lambda I}  +\mbf^\top (G \cdot \diag[\bm{\sigma}] \cdot G^\top + \lambda I)^{-1} \mbf \\
	=& N\log \abs{\lambda} + \sum_{i=1}^n \log(\sigma^i) + \log \abs{\diag[1./\bm{\sigma}] + G^\top G/\lambda} \nonumber \\
	&+ \mbf^\top \left( \frac{1}{\lambda} I - \frac{1}{\lambda^2} G \left(\diag[1./\bm{\sigma}] + \frac{1}{\lambda} G^\top G\right)^{-1} G^\top\right) \mbf \text{,} \nonumber
\end{align}
where the second equation is obtained by applying the Sherman–Morrison–Woodbury formula~\cite{SheMor50} to both summands of $\calE$, and `$1./\bm{\sigma}$' is the element-wise reciprocal of $\bm{\sigma}$.
Since $\frac{1}{\lambda}\mbf^\top \mbf$ and $N\log\abs{\lambda}$ are also independent of $\bm{\sigma}$, minimizing $\calE$ is equivalent to minimizing
\begin{align}
	\calE'(\bm{\sigma}) =& \sum_{i=1}^n \log(\sigma^i) + \log\abs{\diag[1./\bm{\sigma}] + G^\top G/\lambda} \\
	& -\frac{1}{\lambda^2} \mbf^\top \left( G \left(\diag[1./\bm{\sigma}] + \frac{1}{\lambda} G^\top G\right)^{-1} G^\top \right) \mbf \text{.} \nonumber
\end{align}
%
\begin{figure*}[t]
	\centering
	\includegraphics[width=0.98\linewidth]{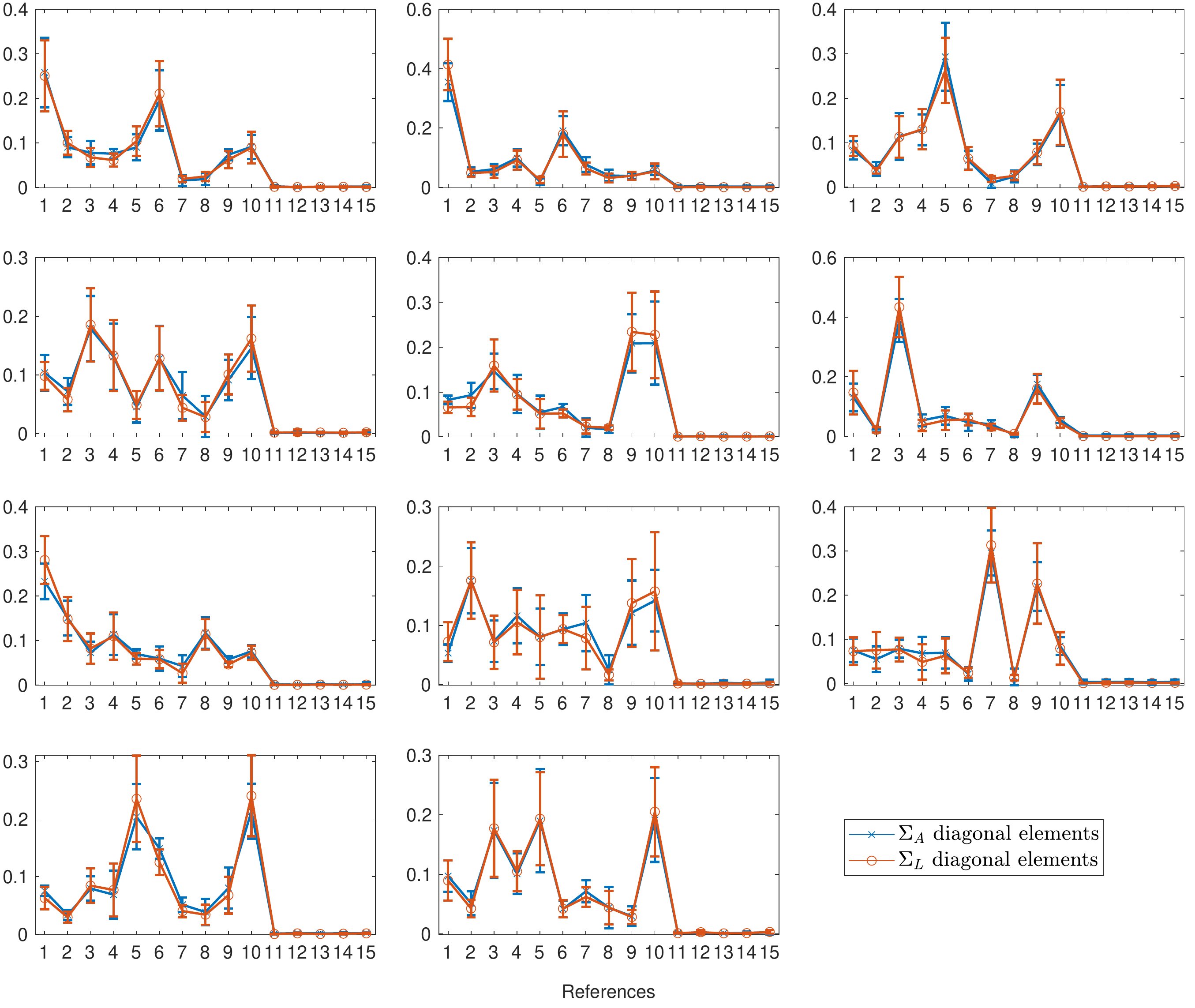}
	\caption{%
		The average diagonal values of $\Sigma_A$ and $\Sigma_L$ optimized for each attribute in the \emph{Pubfig} dataset as the target with the remaining 10 attributes in the same dataset (references 1 to 10), plus 5 additional randomly generated attributes (references 11 to 15) as references.
		The values of $\Sigma_A$ and $\Sigma_L$ are normalized such that the respective sums total to one.
		Note that the unrelated references (11–15) are correctly detected as irrelevant (small magnitudes and standard deviations) and hence ignored.
		In addition, the linear kernel $\Sigma_L$ (orange) is highly correlated to the anisotropic kernel $\Sigma_A$ (blue), so we use a scaled version of $\Sigma_L$ as a surrogate to the optimal $\Sigma_A^*$ (Eq.~\ref{e:kernelscaling}). 
	}
	\label{f:sigmacomp}
\end{figure*}%
Since this energy $\calE'$ is a continuously differentiable function of $\bm{\sigma}$, it can be minimized by standard gradient descent.
Figure~\ref{f:sigmacomp} shows example parameters $\bm{\sigma}^*$ optimized for the \emph{Pubfig} dataset.
For each of the 11 attributes in \emph{Pubfig} as a target, we optimized the corresponding parameters $\bm{\sigma}$ using the remaining attributes as references, plus 5 additional randomly generated references.
As indicated by small magnitudes and the corresponding standard deviations of the $\bm{\sigma}^*$ entries, our algorithm successfully disregards these irrelevant references.
For comparison, we also show the corresponding parameters optimized for the anisotropic Gaussian kernel $k_A$ (Eq.~\ref{e:nonlinearkernel}), demonstrating that once normalized, their relative scaling behaviors are similar, i.e.
\begin{align}
	\label{e:kernelscaling}
	\Sigma_A^* \approx \frac{\Sigma_L^*}{\sigma_k^2}
\end{align}
for a global scaling parameter $\sigma_k^2$.
Our final algorithm uses $\Sigma_L^*/\sigma_k^2$ as a surrogate to $\Sigma_A^*$, using $\sigma_k^2$ as hyperparameter.

\subsection{Algorithm summary}
\label{e:algorithmsummary}

Given the reference matrix $G$, the initial predictor $\mbf^0$, and hyperparameters (noise variance $\sigma^2$ in Eq.~\ref{e:qprime}; global kernel scaling $\sigma^2_k$ in Eq.~\ref{e:kernelscaling}; regularization parameter $\lambda_\text{J}$ in Eq.~\ref{e:objectivef}), our algorithm constructs a denoised predictor by iteratively maximizing the objective
%
\begin{align}
	\label{e:objectivef}
	\calO_\text{N}(\mbf) =& \frac{\mbf^\top A\mbf}{\mbf^\top C_N \mbf} \quad \text{with} \\
	A  =& (C_N \mbf^t)(C_N\mbf^t)^\top + \lambda_\text{J} Q' \quad \text{and} \nonumber\\
	Q' =& C_N (2 K(K + \sigma^2 I)^{-1} - (K + \sigma^2I)^{-1}  K  K(K+\sigma^2I)^{-1}) C_N \text{.}
\label{e:qprime}
\end{align}
%
Algorithm~\ref{a:algorithm} summarizes this process.

\begin{algorithm}[h!]
\caption{Nonlinear 
predictor combination}
\algnewcommand\Input{\item[\textbf{Input:}]}
\algnewcommand\Output{\item[\textbf{Output:}]}
\begin{algorithmic}[1]
    \Input Initial predictor $\mbf^0$, references $\{\mbg_i\}_{i=1}^R$, noise variance $\sigma^2$ (Eq.~\ref{e:qprime}), global kernel scaling $\sigma^2_k$, regularization parameter $\lambda_\text{J}$ (Eq.~\ref{e:objectivef}), and iteration number $S$.
    \STATE $\calH^0=\{\mbh_0^0,\mbh_1^0,\ldots,\mbh_R^0\}\Leftarrow \{\mbf^0,\mbg_1,\ldots,\mbg_R\}$;
    \STATE Calculate the kernel parameter matrix $(\Sigma_{L}^*)_i$ for each $\mbh_i^0\in\calH^0$; $(\Sigma_A)_i=(\Sigma_{L}^*)_i/\sigma^2_k$;
    \FOR{step $t \in \{0, \ldots, S-1\}$}
    \FOR{reference $i \in \{0, \ldots,R\}$}
    \STATE Calculate $\mbh_i^{t+1}\in\calH^{t+1}$ by maximizing $\calO_\text{N}$ based on $\calH^{t}$
    (Eq.~\ref{e:objectivef});
    \ENDFOR
    \ENDFOR
    \Output Denoised target predictor $\mbf^*=\mbh_0^S$.
\end{algorithmic}
\label{a:algorithm}
\end{algorithm}

\section{(Adapting) Existing algorithms}
\label{s:adaptations}

\subsection{Mejjati~\etal's \emph{MTL} algorithm}
\label{s:adaptmtl}

Mejjati~\etal's \emph{MTL} algorithm considers each task-specific predictor as a random variable.
Then, the relationships between tasks are modeled based on the statistical dependence estimated by evaluating these predictor random variables on a dataset $X$~\cite{MejCosKim18}.
Adopting a nonparametric measure of statistical dependence, the finite set independence criterion (FSIC)~\cite{JitSzaGre15}, \emph{MTL} enables training multiple predictors independently of their parametric forms and, therefore, it can be applied to predictor combination problems.

Applying this algorithm to the predictor combination setting, we construct the initial predictor matrix $H^0$ by stacking column-wise, the initial target predictor $\mbf^0$ and the references $\{\mbg_1,\ldots, \mbg_R\}$ 
%
\begin{align}
    H^0 = [\mbf^0, \mbg_1, \ldots, \mbg_R] \text{.}
\end{align}
\emph{MTL} then refines the initial predictor matrix $H^0$ by minimizing the energy
\begin{align}
    \calE_\text{M}(H) =& \norm{H - H^0}^2_\text{F} -\lambda_1 \norm{\vecop(\Phi(H))}_2^2 + \lambda_2 \norm{\vecop(\Phi(H))}_1 \text{,}
\label{e:mtlenergy}
\end{align}
where $\vecop(A)$ constructs a vector by concatenating columns of matrix $A$, $\Phi(H)$ is an $(R+1)\times(R+1)$-sized matrix consisting of pairwise FSIC evaluations: $\Phi(H)_{[i,j]}$ takes a large positive value when $H_{[:,i]}$ and $H_{[:,j]}$ exhibit strong statistical dependence and it takes 0 when $H_{[:,i]}$ and $H_{[:,j]}$ are independent as realizations of random variables.
Minimizing $\calE_\text{M}$ strengthens overall task dependence via (negation of) the $L^2$ norm of $\vecop(\Phi(H))$ and, at the same time, introduces sparsity in the task dependence via the $L^1$ norm of $\Phi(H)$.
Combing these two terms, \emph{MTL} selectively enforces task dependence while suppressing the dependence of weakly related tasks as outliers.
As $\calE_\text{M}$ is not differentiable, standard gradient-descent type algorithms are not applicable.
Instead, it is minimized based on the alternating direction method of multipliers (ADMM) approach.
This involves iteratively solving ADMM sub-problems~\cite{BoyParChu10}, with the number of total iterations $S$ as a hyperparameter.
Once the optimal predictor matrix $H^*$ 
is constructed, the final denoised predictor is obtained by extracting the first column of $H^*$: $\mbf^* = H^*_{[:,1]}$. 
Similarly to our algorithm, $S$ is determined by setting the maximum number of iterations at 50 
and selecting the iteration number achieving the highest validation accuracy.
The other two hyperparameters, $\lambda_1$ and $\lambda_2$, are tuned based on validation accuracy.

\subsection{Derivation of Kim~\etal's algorithm (\emph{OPC}).}
\label{s:adaptkim}

Kim \etal's original predictor combination (\emph{OPC}) approach iteratively minimizes the following energy (Eq.~1 in the main paper):
\begin{align}
    \label{e:ttc}
    \calE_\text{O}(f) &= D_\text{KL}(f \mid f^t)^2 + \lambda_\text{O}\sum_{i=1}^R w_i D_\text{KL}(f \mid g_i)^2 \text{,} \\
    w_i &= \exp\left(-\frac{D_\text{KL}(f^t \mid g_i)^2}{\sigma^2_\text{O}}\right) \text{,}
\end{align}
with $\lambda_\text{O}, \sigma^2_\text{O}>0$ being hyperparameters.
We present how this algorithm is obtained as an instance of Hein and Maier's \emph{Manifold Denoising} algorithm~\cite{HeiMai07} by discretizing a diffusion process on a predictor manifold $\calM$.

\paragraph{Manifold denoising~\cite{HeiMai07}.}
Suppose that we have a set of data points $\calH^0 = \{\mbh^0_i\}_{i=1}^n$ presented as a sample from a Euclidean space $\R^d$ and, further, that the points in $\calH^0$ are sampled from an underlying data-generating manifold $\calM$ embedded in $\R^d$ ($\imath(\calM) \subset \R^d$ with $\imath$ being the embedding), and they are observed as a subset of $\R^d$ contaminated with i.i.d. Gaussian noise $\boldsymbol\epsilon$ in $\R^d$:
\begin{align}
    \mbh_i^0 = \imath(\widetilde{\mbh}_i^0) + \boldsymbol\epsilon \in \R^d \quad \text{for} \quad \widetilde{\mbh}_i^0 \in \calM \text{.}
\end{align}
The manifold denoising algorithm denoises $\calH^0$ by simulating diffusion on a graph $G$ that discretizes $\calM$ (each point $\mbh_i^0 \in \calH^0$ forms a vertex of $G$):
\begin{align}
    \label{e:diff}
    \frac{\partial H}{\partial t}  = - \delta L H \text{,}
\end{align}
where $H = [\mbh_1, \ldots, \mbh_n]^\top $ and $L$ is the graph Laplacian:
\begin{align}
    L &= I - D^{-1} W \text{,} \\
    W_{[i,j]} &= \exp\left(-\frac{\norm{\mbh_i - \mbh_j}^2}{\sigma^2}\right) \text{,}
\label{e:weight}
\end{align}
and $D$ is a diagonal matrix consisting of row sums of $W$, such that $D_{ii} = \sum_{j=1} W_{[i,j]}$.
Now discretizing Eq.~\ref{e:diff} using the implicit Euler method, we obtain
\begin{align}
    H^{t+1} - H^t = -\delta L H^{t+1} \text{.}
    \label{e:disceretediff}
\end{align}
At each time step $t$, the solution $H^{t+1}$ of Eq.~\ref{e:disceretediff} is obtained as the minimizer of the following energy:
\begin{align}
    \label{e:regressiondiff}
    \calE_\text{D}(H) = \norm{H - H^t}_\text{F}^2 + \delta \tr[H^\top L H] \text{,}
\end{align}
where $\norm{A}_\text{F}$ and $\tr[A]$ are the Frobenius norm and trace of matrix $A$, respectively.
As the number of data points $n$ grows to infinity, $G$ becomes a precise representation of $\calM$ embedded in $\R^d$, and 
$L$ converges to the Laplace-Beltrami operator $\Delta_\calM$ on $\calM$ casting Eq.~\ref{e:disceretediff} into a diffusion process on a continuous manifold $\calM$~\cite{HeiMai07}.
It should be noted that the graph Laplacian was constructed based on the \emph{ambient} $L^2$ distance in $\R^d$ rather than the intrinsic metric on $\calM$.
This facilitates building a practical, still consistent algorithm: Equation~\ref{e:regressiondiff} only requires the ambient Euclidean distance (via $L$) without having to access $\calM$ directly, but it guarantees the statistical consistency of $L$ as proven by Hein and Maier~\cite{HeiMai07}.

Now applying this algorithm to the predictor combination setting and, therefore, assuming that only the first point $\mbh_1^0 \in \calH^0$ is noisy (\ie $\mbh_i = \imath(\tilde{\mbh}_i)$ for $i = \{2, \ldots, n\}$), we obtain an iterative update rule of $\mbh_1^t$ given fixed \emph{references} $\{\mbh_i\}_{i=2}^n$:
\begin{align}
    \label{e:regressionforpc}
    \calE_\text{D}(\mbh) = \norm{\mbh - \mbh_1^t}^2 + \delta\sum_{i=2}^n W_{[1,i]} \norm{\mbh - \mbh_i}^2 \text{.}
\end{align}
%
Finally, $\calE_\text{O}$ in Eq.~\ref{e:ttc} is obtained by replacing each point in $\calH^t$ and the corresponding $L^2$ distances in $\calE_\text{D}$ with a Gaussian process predictor and Kullback–Leibler divergences, respectively: $\mbh_1^t$ and $\{\mbh_i\}_{i=2}^n$ (with $n=R+1$) 
are considered as the target predictor $\mbf$ and the corresponding references $\{\mbg_i\}$, respectively.

\subsection{Evgeniou~\etal's graph Laplacian (\emph{GL})-based MTL algorithm.}
\label{s:adaptmtl2}


Evgeniou~\etal's graph Laplacian (\emph{GL})-based algorithm learns predictors $\calH = \{h_i\}_{i=1}^{n}$ of multiple tasks by enforcing pairwise parameter similarities: Assuming that all predictors are linear, \ie $h_i(\mbx) = \mbw_i^\top \mbx$, their algorithm estimates the predictor parameters $W = \{\mbw_1, \ldots, \mbw_{n}\}$ by minimizing the energy
\begin{align}
    \label{e:glcost}
    \calE_\text{GL}(W) =& \sum_{i=1}^n l_i(h_i) + \lambda_1 \sum_{i=1}^n \norm{\mbw_i}^2 +\lambda_2\sum_{i=1}^n\sum_{j\neq i} U_{[i,j]} \norm{\mbw_i - \mbw_j}^2 \text{,}
\end{align}
where $\{l_i(\cdot)\}_{i=1}^n$ are task-specific loss functions and $U_{[i,j]} \geq 0$ represents the relationship between tasks $i$ and $j$.
Now adapting this algorithm to the predictor combination setting, we assume that the initial predictor $f^0 = h_1$ and the corresponding references $g_i = h_{i+1}$ for $i \in \{1, \ldots, R\}$ are given ($n = R+1$).
Then, $f^0$ is refined by minimizing the energy
\begin{align}
    \label{e:gladaptation}
    \calE_\text{GL}(\mbw) &= \norm{\mbw - \mbw_1^0}^2 + \lambda_\text{GL}\sum_{j=2}^n U_{[1,j]} \norm{\mbw - \mbw_j}^2 \text{.}
\end{align}
In general, determining the task relationship parameters $\{U_{[1,j]}\}$ is a challenging problem.
Here, we determine them by adopting Kim~\etal's approach: We iteratively update $\{U_{[1,j]}\}$ by minimizing $\calE_\text{GL}$ at each time step $t$ with
\begin{align}
    U_{[1,j]} &= \exp\left(-\frac{\norm{\mbw_i - \mbw_1^t}^2}{\sigma_\text{GL}^2}\right) \text{.}
    \label{e:glweight}
\end{align}
Further, adopting Kim and Chang's approach~\cite{KimCha19}, we explicitly constrain all predictor parameter vectors to have unit norm: $\norm{\mbw_i} = 1$, enabling the comparison of task predictor parameters independently of their scales.
The two hyperparameters $\sigma_\text{GL}^2$ and $\lambda_\text{GL}$ are determined based on validation accuracy.
As often, ranking problems are nonlinear, we extend this framework by adopting  
the linear-in-parameter model:
\begin{align}
    f(\mbx) = \phi(\mbx)^\top \mbw_f \text{,}
\end{align}
where $\phi \colon \calX \to \calF_k$ with $\calF_k$ being the reproducing kernel Hilbert space (RKHS) corresponding to a Gaussian kernel with hyperparameter $\sigma_k^2$~\cite{Sch02}:
\begin{align}
    k(\mba,\mbb) = \exp\left(-\frac{\norm{\mba - \mbb}^2}{\sigma_k^2}\right) \text{.}
\end{align}
In this case, the target predictor $f$ is represented based the original parameter vector $\mbw_f$ as well as its dual parameter vector $\mba_f = [a^1_f, \ldots, a^{N'}_f]^\top$:
\begin{align}
    \label{e:kernelspanf}
    f(\mbx) := \phi(\mbx)^\top \mbw_f = \sum_{j=1}^{N'} a_f^j k(\mbb_j, \mbx) \text{,}
\end{align}
%
with $\{\mbb_i\}_{i=1}^{N'}$ being a set of \emph{basis vectors}.
The reference predictors $\{g_i\}_{i=1}^R$ are represented similarly:
%
\begin{align}
    \label{e:kernelspang}
    g_i(\mbx) := \phi(\mbx)^\top \mbw_i = \sum_{j=1}^{N'} a_i^j k(\mbb_j, \mbx) \text{.}
\end{align}
Under this setting, the parameter similarity $\norm{\mbw_f - \mbw_i}$ can be calculated using the standard kernel trick~\cite{Sch02} as
\begin{align}
    \norm{\mbw_f - \mbw_j} = \mba_f^\top K \mba_f^\top + \mba_i^\top K \mba_i^\top - 2\mba_f^\top K \mba_i^\top \text{,}
    \label{e:kernelpdistance}
\end{align}
with $K_{[i,j]} = k(\mbb_i, \mbb_j)$.
It should be noted that efficient\footnote{The RKHS $\calF_k$ corresponding to a Gaussian kernel $k$ is infinite-dimensional. Therefore, each parameter vector $\mbw\in \calF_k$ is an infinite-dimensional object, making the direct evaluation of $\|\mbw_f-\mbw_j\|$ infeasible.} calculation of $\|\mbw_f-\mbw_i\|$ based on Eq.~\ref{e:kernelpdistance} requires that all predictors should share the same RKHS determined by the kernel parameter $\sigma_k$.
To facilitate this, in our experiments, we first determine $f^0$ as nonlinear rank support vector machine that minimizes the regularized energy
\begin{align}
\calE_\text{S}(f) &= \sum_{(\mbx_i, \mbx_j) \in U} \calL(f, (\mbx_i, \mbx_j)) + C_f \|\mbw\|^2 \text{,} \\
\calL(f, (\mba, \mbb)) &= \max\left(1 - (f(\mba) - f(\mbb)), 0\right)^2 
\label{e:rankloss}
\end{align}
for the rank loss $\calL$ defined on ground-truth ranked pairs $U\subset X\times X$  
%
and tune the hyperparameters $\sigma^2_{k}$ and $C_f$ based on validation accuracy.
Once $f^0$ is fixed in this way, the reference predictors $\{g_i\}_{i=1}^R$ are determined by minimizing $\calE_\text{S}$ for the respective rank labels.
However, for these references, only the respective regularization hyperparameters $\{C_i\}_{i=1}^R$ are tuned while the corresponding kernel parameters are all fixed as $\sigma^2_{k}$ (optimized for $f^0$), to facilitate the computation of $\norm{\mbw - \mbw_i}$ (Eq.~\ref{e:kernelpdistance}).
We fixed $N'$ at 500 and selected the basis vectors $\{\mbb_i\}_{i=1}^{N'}$ as the cluster centers of input data points $X$, estimated using $k$-means clustering.

Note that this setting violates the application conditions of predictor combination: It requires access to the forms of all predictors $\{f, g_1, \ldots, g_R\}$ and, further, it assumes that all predictors share the same form (Eqs.~\ref{e:kernelspanf} and \ref{e:kernelspang}).
We show in Sec.~\ref{s:experiment} that the latter homogeneity requirement poses a severe limitation on predictor combination performance.
Even when \emph{GL} took advantage of known predictor forms, except for a few cases, the other predictor combination algorithms significantly outperformed $\emph{GL}$.
Often, the results of $\emph{GL}$ are even worse than the initial predictors $\mbf^0$ that are obtained by selecting the best predictors (via validation) from the heterogeneous predictor pools.

\section{Complete ranking results}
\label{s:experiment}


\noindent
Table~\ref{t:significancetest} summarizes the results for the relative attributes ranking experiments (see Tables~\ref{t:rankingresultsshoes}--\ref{t:awa2results2} for complete results).
Our algorithm \emph{NPC} performs best for 87\% (162/186) of attributes.
In particular, it showed statistically significant improvement on 74 out of 80 \emph{AWA2} attributes, while the baselines \emph{OPC} and \emph{MTL} achieved significant performance gains only on 8 and 33 attributes, respectively.%
\footnote{%
	We used a t-test with $\alpha=0.95$.
	Note that statistical significance tests do not necessarily evaluate how significant the improvements are in an absolute scale:
	Even when the improvements are marginal, if they are consistent, the result of statistical significant tests can be positive.
	For instance, for \emph{AWA2} attribute 4, \emph{MTL} achieved rather moderate improvements (with mean 0.05) but the test of statistical significance is positive as the results consistently improved the performance from the baseline, as indicated by the small standard deviation.}

Overall, our algorithm \emph{NPC} is often statistically significantly better than these methods and – apart from only one attribute (for \emph{CUB}) out of 186 – ours is not statistically significantly worse than the other methods.
This demonstrates that the baselines \emph{OPC} and \emph{MTL} are limited in that they can only capture pairwise dependence between the target predictor and each reference.
Taking into account the dependence present among the references, and thereby \emph{jointly} exploiting them in improving the target predictor, our algorithm \emph{NPC (Ours)} 
significantly improves the performance. 

\begin{table}[t]
\caption{\label{t:significancetest}%
A summary of the results of statistical significance tests of our method NPC compared to baseline $\mbf^0$, \emph{GL}, \emph{OPC} and \emph{MTL}, based on a t--test with $\alpha = 0.95$.
For each method, we show \#attributes where our \emph{NPC} is statistically significantly better (\BET{first column}), on par with (\PAR{second column}), and statistically significantly worse (\WOR{third column}). 
}
\centering
\resizebox{\linewidth}{!}{%
\renewcommand*{\arraystretch}{1}
\setlength{\tabcolsep}{6pt}%
\begin{tabular}{l|rrr|rrr|rrr|rrr|r}
\toprule
Dataset&\multicolumn{3}{c|}{vs. baseline $\mbf^0$}& \multicolumn{3}{c|}{vs. \emph{GL}}& \multicolumn{3}{c|}{vs. \emph{OPC}}& \multicolumn{3}{c|}{vs. \emph{MTL}}& \# total attr.\\
\midrule
\emph{Shoes}&\BET{9}&\PAR{1}&\WOR{0}&\BET{10}&\PAR{0}&\WOR{0}&\BET{8}&\PAR{2}&\WOR{0}&\BET{9}&\PAR{1}&\WOR{0}&10\\
\midrule
\emph{Pubfig}&\BET{11}&\PAR{0}&\WOR{0}&\BET{11}&\PAR{0}&\WOR{0}&\BET{11}&\PAR{0}&\WOR{0}&\BET{11}&\PAR{0}&\WOR{0}&11\\
\midrule
\emph{OSR}&\BET{6}&\PAR{0}&\WOR{0}&\BET{2}&\PAR{4}&\WOR{0}&\BET{4}&\PAR{2}&\WOR{0}&\BET{5}&\PAR{1}&\WOR{0}&6\\
\midrule
\emph{OSR (ResNet)}&\BET{6}&\PAR{0}&\WOR{0}&\BET{6}&\PAR{0}&\WOR{0}&\BET{6}&\PAR{0}&\WOR{0}&\BET{6}&\PAR{0}&\WOR{0}&6\\
\midrule
\emph{aPascal}&\BET{25}&\PAR{4}&\WOR{0} &\BET{16}&\PAR{13}&\WOR{0}&\BET{9}&\PAR{20}&\WOR{0}&\BET{19}&\PAR{10}&\WOR{0}&29\\
\midrule
\emph{CUB}&\BET{31}&\PAR{9}&\WOR{0}&\BET{23}&\PAR{17}&\WOR{0}&\BET{26}&\PAR{14}&\WOR{0}&\BET{12}&\PAR{27}&\WOR{1}&40\\
\midrule
\emph{AWA2}&\BET{74}&\PAR{6}&\WOR{0}&\BET{73}&\PAR{7}&\WOR{0}&\BET{71}&\PAR{9}&\WOR{0}&\BET{72}&\PAR{8}&\WOR{0}&80\\
\midrule
\emph{Zap50K}&\BET{0}&\PAR{4}&\WOR{0}&\BET{0}&\PAR{4}&\WOR{0}&\BET{0}&\PAR{4}&\WOR{0}&\BET{0}&\PAR{4}&\WOR{0}&4\\
\midrule\midrule
\# total attr.&\BET{162}&\PAR{24}&\WOR{0}&\BET{141}&\PAR{45}&\WOR{0}&\BET{135}&\PAR{51}&\WOR{0}&\BET{134}&\PAR{51}&\WOR{1}&186\\
\bottomrule
\end{tabular}%
}
\end{table}

%


\begin{table*}[p]
\caption{\label{t:rankingresultsshoes}%
    Ranking accuracies of different predictor combination algorithms on the \emph{Shoes}, \emph{Pubfig}, \emph{OSR}, and \emph{OSR (ResNet)} datasets. For each dataset, we repeated experiments 10 times with different training, validation, and test 
set splits.  
    For baseline $\mbf^0$ (second column), Kendall's Tau correlations$\times$100 (standard deviations in parentheses) are presented.
    For the remaining algorithms (third to sixth columns), the accuracy offsets from $\mbf^0$ are presented.
    The best and second best results are highlighted with \ROne{bold} and \RTwo{italic} fonts, respectively. The results of statistical significance test based on a t--test with $\alpha = 0.95$ are highlighted in \PST{green} (significantly positive) and \NST{orange} (significantly negative). The last three columns show the results of statistical significance test of our algorithm with \emph{GL}, \emph{OPC}, and \emph{MTL}, respectively
    ($\PST{+}$/$\NST{-}$: significantly positive/negative).
}
\centering
\resizebox{\linewidth}{!}{%
\renewcommand*{\arraystretch}{0.88}
\setlength{\tabcolsep}{3pt}%
\begin{tabular}{crrrrr|ccc}
\toprule
\multicolumn{9}{c}{\emph{Shoes}}\\
\midrule
Attr. & Baseline $\mbf^0$ & \emph{GL} & \emph{OPC} & \emph{MTL} & \emph{NPC (ours)}& vs. \emph{GL} & vs. \emph{OPC} & vs. \emph{MTL} \\
\midrule
1 & 72.09 (1.71)  & -0.36 (1.44)& \PST{\RTwo{2.36 (0.79)}}& \PST{2.03 (0.49)}& \PST{\ROne{3.21 (0.86)}}& $\PST{+}$& $\PST{+}$& $\PST{+}$\\
2 & 63.84 (1.87)  & -2.04 (2.94)& \PST{\RTwo{1.57 (1.25)}}& \PST{0.70 (0.39)}& \PST{\ROne{2.26 (1.38)}}& $\PST{+}$& $0$& $\PST{+}$\\
3 & 38.07 (2.11)  & -1.38 (2.43)& -0.24 (0.65)& \RTwo{0.16 (0.70)}& \PST{\ROne{4.58 (2.45)}}& $\PST{+}$& $\PST{+}$& $\PST{+}$\\
4 & \RTwo{50.10 (2.75)} & \NST{-2.45 (2.59)}& -0.88 (3.29)& -0.08 (0.51)& \PST{\ROne{1.63 (2.25)}}& $\PST{+}$& $\PST{+}$& $\PST{+}$\\
5 & 65.76 (1.20)  & -1.11 (1.84)& -0.05 (0.16)& \RTwo{0.13 (0.34)}& \ROne{0.92 (2.46)}& $\PST{+}$& $0$& $0$\\
6 & 65.02 (1.83)  & \NST{-0.86 (1.13)}& \PST{0.68 (0.87)}& \PST{\RTwo{0.81 (0.86)}}& \PST{\ROne{4.18 (1.54)}}& $\PST{+}$& $\PST{+}$& $\PST{+}$\\
7 & 59.38 (2.06)  & \NST{-3.31 (1.59)}& \RTwo{0.78 (1.16)}& \PST{0.45 (0.42)}& \PST{\ROne{4.14 (3.20)}}& $\PST{+}$& $\PST{+}$& $\PST{+}$\\
8 & 56.85 (2.04)  & \NST{-2.57 (1.46)}& 0.19 (0.46)& \PST{\RTwo{0.40 (0.51)}}& \PST{\ROne{2.62 (0.87)}}& $\PST{+}$& $\PST{+}$& $\PST{+}$\\
9 & 65.15 (1.94)  & 0.35 (1.94)& \PST{\RTwo{2.49 (1.27)}}& \PST{1.40 (0.72)}& \PST{\ROne{4.58 (1.72)}}& $\PST{+}$& $\PST{+}$& $\PST{+}$\\
10 & 72.10 (1.24)  & \NST{-1.26 (1.55)}& \PST{\RTwo{1.71 (0.77)}}& \PST{1.47 (0.77)}& \PST{\ROne{2.75 (1.05)}}& $\PST{+}$& $\PST{+}$& $\PST{+}$\\
\bottomrule
\toprule
\multicolumn{9}{c}{\emph{Pubfig}}\\
\midrule
Attr. & Baseline $\mbf^0$ & \emph{GL} & \emph{OPC} & \emph{MTL} & \emph{NPC (ours)}& vs. \emph{GL} & vs. \emph{OPC} & vs. \emph{MTL} \\
\midrule
1 & 67.13 (2.75)  & \PST{4.89 (2.48)}& \PST{8.37 (3.84)}& \PST{\RTwo{9.37 (2.94)}}& \PST{\ROne{15.45 (2.59)}}& $\PST{+}$& $\PST{+}$& $\PST{+}$\\
2 & 62.49 (2.41)  & -0.96 (2.62)& -0.31 (0.91)& \PST{\RTwo{2.24 (1.55)}}& \PST{\ROne{13.78 (3.23)}}& $\PST{+}$& $\PST{+}$& $\PST{+}$\\
3 & 68.31 (2.33)  & 2.27 (3.27)& \PST{3.06 (2.52)}& \PST{\RTwo{6.25 (3.04)}}& \PST{\ROne{11.33 (3.10)}}& $\PST{+}$& $\PST{+}$& $\PST{+}$\\
4 & 63.98 (3.46)  & \PST{\RTwo{8.44 (4.26)}}& \PST{4.88 (3.14)}& \PST{7.84 (3.49)}& \PST{\ROne{17.80 (4.26)}}& $\PST{+}$& $\PST{+}$& $\PST{+}$\\
5 & 61.27 (2.96)  & \PST{\RTwo{6.15 (2.92)}}& \PST{3.33 (4.23)}& \PST{3.26 (3.70)}& \PST{\ROne{16.62 (6.02)}}& $\PST{+}$& $\PST{+}$& $\PST{+}$\\
6 & 81.60 (1.26)  & -1.09 (2.67)& -0.03 (1.25)& \RTwo{0.44 (1.58)}& \PST{\ROne{6.17 (2.03)}}& $\PST{+}$& $\PST{+}$& $\PST{+}$\\
7 & 64.23 (2.88)  & \PST{2.87 (3.13)}& 1.68 (2.67)& \RTwo{3.14 (4.61)}& \PST{\ROne{15.66 (3.14)}}& $\PST{+}$& $\PST{+}$& $\PST{+}$\\
8 & 66.10 (3.53)  & \RTwo{0.38 (2.66)}& \PST{0.19 (0.21)}& 0.10 (0.49)& \PST{\ROne{12.16 (3.17)}}& $\PST{+}$& $\PST{+}$& $\PST{+}$\\
9 & 59.73 (4.79)  & \PST{\RTwo{3.58 (3.78)}}& -0.11 (1.89)& 1.96 (3.20)& \PST{\ROne{17.74 (4.68)}}& $\PST{+}$& $\PST{+}$& $\PST{+}$\\
10 & 63.58 (3.48)  & \PST{\RTwo{5.79 (3.55)}}& \PST{3.24 (1.92)}& \PST{4.06 (2.23)}& \PST{\ROne{14.16 (2.58)}}& $\PST{+}$& $\PST{+}$& $\PST{+}$\\
11 & 69.12 (2.87)  & \PST{7.76 (2.73)}& \PST{9.30 (2.48)}& \PST{\RTwo{9.49 (2.73)}}& \PST{\ROne{15.20 (2.87)}}& $\PST{+}$& $\PST{+}$& $\PST{+}$\\
\bottomrule
\toprule
\multicolumn{9}{c}{\emph{OSR}}\\
\midrule
Attr. & Baseline $\mbf^0$ & \emph{GL} & \emph{OPC} & \emph{MTL} & \emph{NPC (ours)}& vs. \emph{GL} & vs. \emph{OPC} & vs. \emph{MTL} \\
\midrule
1 & 88.57 (0.93)  & \PST{\ROne{3.16 (1.07)}}& \PST{2.06 (0.83)}& \PST{2.19 (1.03)}& \PST{\RTwo{2.72 (1.39)}}& $0$& $0$& $0$\\
2 & 87.52 (0.89)  & \NST{-1.17 (1.06)}& -0.00 (0.19)& \RTwo{0.03 (0.12)}& \PST{\ROne{0.93 (0.69)}}& $\PST{+}$& $\PST{+}$& $\PST{+}$\\
3 & 76.12 (0.95)  & 0.50 (1.32)& \PST{1.31 (0.92)}& \PST{\RTwo{1.99 (1.26)}}& \PST{\ROne{3.25 (1.49)}}& $\PST{+}$& $\PST{+}$& $\PST{+}$\\
4 & 77.67 (0.92)  & \PST{\RTwo{1.29 (1.11)}}& \PST{0.69 (0.70)}& \PST{0.90 (0.68)}& \PST{\ROne{2.10 (1.29)}}& $0$& $\PST{+}$& $\PST{+}$\\
5 & 79.58 (0.65)  & \PST{\RTwo{2.50 (0.72)}}& \PST{2.26 (0.68)}& \PST{1.43 (0.86)}& \PST{\ROne{2.89 (1.04)}}& $0$& $0$& $\PST{+}$\\
6 & 80.49 (1.22)  & \RTwo{0.70 (1.00)}& 0.09 (0.52)& 0.03 (0.43)& \PST{\ROne{1.46 (0.84)}}& $0$& $\PST{+}$& $\PST{+}$\\
\bottomrule
\toprule
\multicolumn{9}{c}{\emph{OSR (ResNet)}}\\
\midrule
Attr. & Baseline $\mbf^0$ & \emph{GL} & \emph{OPC} & \emph{MTL} & \emph{NPC (ours)}& vs. \emph{GL} & vs. \emph{OPC} & vs. \emph{MTL} \\
\midrule
1 & 96.12 (0.59)  & -0.07 (0.14)& \PST{\RTwo{0.39 (0.33)}}& 0.26 (0.37)& \PST{\ROne{1.33 (0.48)}}& $\PST{+}$& $\PST{+}$& $\PST{+}$\\
2 & 84.73 (0.87)  & 0.00 (0.27)& -0.11 (0.25)& \RTwo{0.10 (0.25)}& \PST{\ROne{2.51 (1.14)}}& $\PST{+}$& $\PST{+}$& $\PST{+}$\\
3 & 84.46 (1.08)  & -0.01 (0.03)& \PST{0.43 (0.30)}& \PST{\RTwo{0.89 (0.66)}}& \PST{\ROne{2.56 (1.21)}}& $\PST{+}$& $\PST{+}$& $\PST{+}$\\
4 & 85.14 (1.27)  & -0.09 (0.44)& -0.03 (0.10)& \RTwo{0.57 (0.89)}& \PST{\ROne{2.45 (0.78)}}& $\PST{+}$& $\PST{+}$& $\PST{+}$\\
5 & 88.00 (0.78)  & 0.03 (0.24)& 0.55 (0.82)& \RTwo{0.62 (0.93)}& \PST{\ROne{3.52 (1.66)}}& $\PST{+}$& $\PST{+}$& $\PST{+}$\\
6 & 90.88 (0.88)  & -0.08 (0.17)& 0.06 (0.25)& \PST{\RTwo{0.71 (0.57)}}& \PST{\ROne{1.56 (1.08)}}& $\PST{+}$& $\PST{+}$& $\PST{+}$\\
\bottomrule
\end{tabular}%
}
\end{table*}

\begin{table*}[p]
\caption{\label{t:rankingresultsapascal}%
    Ranking accuracies of different predictor combination algorithms on the \emph{aPascal} and \emph{Zap50K} datasets. For each dataset, we repeated experiments 10 times with different training, validation, and test 
set splits.  
    For baseline $\mbf^0$ (second column), Kendall's Tau correlations$\times$100 (standard deviations in parentheses) are presented.
    For the remaining algorithms (third to sixth columns), the accuracy offsets from $\mbf^0$ are presented.
    The best and second best results are highlighted with \ROne{bold} and \RTwo{italic} fonts, respectively. The results of statistical significance test based on a t--test with $\alpha = 0.95$ are highlighted in \PST{green} (significantly positive) and \NST{orange} (significantly negative). The last three columns show the results of statistical significance test of our algorithm with \emph{GL}, \emph{OPC}, and \emph{MTL}, respectively
	($\PST{+}$/$\NST{-}$: significantly positive/negative).
}
\centering
\resizebox{\linewidth}{!}{%
\renewcommand*{\arraystretch}{0.88}
\setlength{\tabcolsep}{3pt}%
\begin{tabular}{crrrrr|ccc}
\toprule
\multicolumn{9}{c}{\emph{aPascal}}\\
\midrule
Attr. & Baseline $\mbf^0$ & \emph{GL} & \emph{OPC} & \emph{MTL} & \emph{NPC (ours)}& vs. \emph{GL} & vs. \emph{OPC} & vs. \emph{MTL} \\
\midrule
1 & 59.44 (4.19)  & \PST{3.11 (2.88)}& \PST{\RTwo{3.38 (2.31)}}& \PST{0.87 (0.64)}& \PST{\ROne{6.04 (3.43)}}& $\PST{+}$& $\PST{+}$& $\PST{+}$\\
2 & 68.21 (4.50)  & -0.02 (0.12)& \RTwo{0.42 (0.87)}& \PST{0.27 (0.13)}& \ROne{1.19 (2.17)}& $0$& $0$& $0$\\
3 & 14.45 (4.62)  & 1.15 (2.40)& \RTwo{2.09 (5.79)}& \PST{0.69 (0.33)}& \PST{\ROne{4.24 (4.67)}}& $0$& $0$& $\PST{+}$\\
4 & 65.20 (3.20)  & 0.23 (0.57)& \PST{\ROne{1.82 (1.38)}}& \PST{0.07 (0.04)}& \RTwo{1.19 (2.67)}& $0$& $0$& $0$\\
5 & 57.87 (4.38)  & \PST{2.08 (1.69)}& \PST{\RTwo{4.10 (2.03)}}& \PST{1.19 (0.67)}& \PST{\ROne{4.10 (2.26)}}& $\PST{+}$& $0$& $\PST{+}$\\
6 & 57.37 (5.76)  & \PST{3.21 (1.49)}& \PST{\ROne{4.18 (1.23)}}& \PST{1.36 (0.70)}& \PST{\RTwo{3.90 (2.28)}}& $0$& $0$& $\PST{+}$\\
7 & 71.34 (2.69)  & 0.05 (0.36)& \PST{\RTwo{2.29 (2.27)}}& \PST{1.05 (0.53)}& \PST{\ROne{2.91 (1.85)}}& $\PST{+}$& $0$& $\PST{+}$\\
8 & 67.08 (4.56)  & \PST{2.25 (1.79)}& \PST{\RTwo{3.78 (3.30)}}& \PST{1.65 (0.51)}& \PST{\ROne{3.79 (3.11)}}& $\PST{+}$& $0$& $\PST{+}$\\
9 & 62.36 (4.83)  & \PST{2.54 (1.51)}& \PST{\RTwo{3.91 (2.27)}}& \PST{2.03 (0.55)}& \PST{\ROne{4.69 (1.34)}}& $\PST{+}$& $0$& $\PST{+}$\\
10 & 57.09 (4.36)  & \PST{4.79 (2.25)}& \PST{\RTwo{5.38 (2.86)}}& \PST{2.92 (1.55)}& \PST{\ROne{7.00 (2.74)}}& $\PST{+}$& $\PST{+}$& $\PST{+}$\\
11 & 62.25 (3.59)  & \PST{2.05 (1.54)}& \PST{\RTwo{2.61 (1.39)}}& \PST{1.76 (0.75)}& \PST{\ROne{4.12 (3.01)}}& $0$& $0$& $\PST{+}$\\
12 & 60.58 (4.76)  & \PST{\RTwo{3.48 (2.13)}}& \PST{\ROne{4.25 (2.38)}}& \PST{1.54 (0.67)}& \PST{3.43 (2.63)}& $0$& $0$& $\PST{+}$\\
13 & 46.71 (4.95)  & \PST{3.41 (3.54)}& \PST{\RTwo{4.20 (4.16)}}& \PST{1.47 (0.60)}& \PST{\ROne{4.86 (3.99)}}& $0$& $0$& $\PST{+}$\\
14 & 52.31 (3.52)  & \PST{3.52 (2.89)}& \PST{\RTwo{4.37 (1.68)}}& \PST{2.29 (1.07)}& \PST{\ROne{6.75 (2.27)}}& $\PST{+}$& $\PST{+}$& $\PST{+}$\\
15 & 52.07 (6.42)  & \PST{2.14 (2.77)}& \PST{\RTwo{2.34 (2.77)}}& \PST{1.46 (0.77)}& \PST{\ROne{5.65 (3.71)}}& $\PST{+}$& $\PST{+}$& $\PST{+}$\\
16 & 47.33 (4.15)  & 0.46 (0.72)& \PST{\RTwo{1.55 (1.22)}}& \PST{1.16 (0.41)}& \PST{\ROne{2.76 (2.55)}}& $\PST{+}$& $0$& $0$\\
17 & 49.53 (4.57)  & -0.18 (0.96)& 0.42 (1.70)& \PST{\RTwo{0.95 (0.45)}}& \PST{\ROne{2.49 (2.44)}}& $\PST{+}$& $0$& $0$\\
18 & 82.22 (3.20)  & \PST{0.60 (0.81)}& \PST{\RTwo{1.37 (0.70)}}& \PST{1.10 (0.36)}& \PST{\ROne{1.53 (0.88)}}& $\PST{+}$& $0$& $0$\\
19 & 70.22 (3.21)  & \PST{1.04 (1.05)}& \PST{1.01 (1.37)}& \PST{\RTwo{1.42 (0.61)}}& \PST{\ROne{2.40 (2.19)}}& $0$& $0$& $0$\\
20 & 79.62 (2.60)  & \PST{1.87 (0.88)}& \PST{\ROne{2.10 (1.19)}}& \PST{1.44 (0.58)}& \PST{\RTwo{2.09 (2.31)}}& $0$& $0$& $0$\\
21 & 53.01 (5.18)  & \RTwo{0.69 (1.34)}& 0.02 (0.37)& \PST{0.37 (0.25)}& \PST{\ROne{3.70 (4.07)}}& $\PST{+}$& $\PST{+}$& $\PST{+}$\\
22 & 55.52 (3.98)  & \PST{2.83 (2.30)}& \PST{\RTwo{3.04 (1.78)}}& \PST{2.22 (0.71)}& \PST{\ROne{6.41 (3.92)}}& $\PST{+}$& $\PST{+}$& $\PST{+}$\\
23 & 70.53 (5.23)  & \PST{2.31 (2.23)}& \PST{2.03 (1.14)}& \PST{\RTwo{2.34 (1.11)}}& \PST{\ROne{3.82 (2.24)}}& $\PST{+}$& $\PST{+}$& $\PST{+}$\\
24 & 38.48 (5.51)  & \RTwo{1.12 (2.63)}& 0.95 (3.22)& \PST{0.91 (0.43)}& \ROne{1.81 (4.71)}& $0$& $0$& $0$\\
25 & 48.45 (3.52)  & \RTwo{0.89 (2.29)}& 0.76 (1.65)& \PST{0.58 (0.51)}& \PST{\ROne{2.48 (2.88)}}& $0$& $\PST{+}$& $\PST{+}$\\
26 & 49.93 (3.91)  & \PST{2.53 (2.56)}& \PST{\RTwo{3.60 (3.00)}}& \PST{2.31 (0.55)}& \PST{\ROne{6.00 (2.57)}}& $\PST{+}$& $\PST{+}$& $\PST{+}$\\
27 & 72.87 (3.05)  & 0.04 (0.31)& \PST{\RTwo{1.07 (1.49)}}& \PST{0.91 (0.29)}& \PST{\ROne{1.65 (1.55)}}& $\PST{+}$& $0$& $0$\\
28 & 64.35 (2.27)  & 0.30 (0.66)& \PST{\ROne{1.39 (1.33)}}& \PST{0.42 (0.42)}& \RTwo{0.43 (3.09)}& $0$& $0$& $0$\\
29 & 53.84 (3.47)  & \PST{3.97 (2.36)}& \PST{\RTwo{4.16 (2.19)}}& \PST{2.53 (0.78)}& \PST{\ROne{5.17 (2.20)}}& $0$& $0$& $\PST{+}$\\
\bottomrule
\toprule
\multicolumn{9}{c}{\emph{Zap50K}}\\
\midrule
Attr. & Baseline $\mbf^0$ & \emph{GL} & \emph{OPC} & \emph{MTL} & \emph{NPC (ours)}& vs. \emph{GL} & vs. \emph{OPC} & vs. \emph{MTL} \\
\midrule
1 & 87.97 (0.99)  & -0.00 (0.38)& \ROne{0.27 (0.60)}& -0.20 (0.67)& \RTwo{0.27 (0.75)}& $0$& $0$& $0$\\
2 & 89.43 (1.56)  & \RTwo{0.13 (0.82)}& -0.27 (1.10)& \ROne{0.40 (0.86)}& 0.03 (0.95)& $0$& $0$& $0$\\
3 & 90.67 (1.29)  & 0.43 (0.80)& \PST{\ROne{0.80 (0.83)}}& \PST{\RTwo{0.70 (0.84)}}& 0.67 (1.23)& $0$& $0$& $0$\\
4 & 90.33 (1.56)  & \RTwo{0.23 (1.14)}& 0.17 (0.98)& 0.03 (0.82)& \ROne{0.37 (0.87)}& $0$& $0$& $0$\\
\bottomrule
\end{tabular}%
}
\end{table*}

\begin{table*}[p]
\caption{\label{t:rankingresultscub}%
    Ranking accuracies of different predictor combination algorithms on the \emph{CUB} dataset. We repeated experiments 10 times with different training, validation, and test set splits. For baseline $\mbf^0$ (second column), Kendall's Tau correlations$\times$100 (standard deviations in parentheses) are presented. For the remaining algorithms (third to sixth columns), the accuracy offsets from $\mbf^0$ are presented.
    The best and second best results are highlighted with \ROne{bold} and \RTwo{italic} fonts, respectively. The results of statistical significance test based on a t--test with $\alpha = 0.95$ are highlighted in \PST{green} (significantly positive) and \NST{orange} (significantly negative). The last three columns show the results of statistical significance test of our algorithm with \emph{GL}, \emph{OPC}, and \emph{MTL}, respectively
	($\PST{+}$/$\NST{-}$: significantly positive/negative).
}
\centering
\resizebox{\linewidth}{!}{%
\renewcommand*{\arraystretch}{0.88}
\setlength{\tabcolsep}{3pt}%
\begin{tabular}{crrrrr|ccc}
\toprule
Attr. & Baseline $\mbf^0$ & \emph{GL} & \emph{OPC} & \emph{MTL} & \emph{NPC (ours)}& vs. \emph{GL} & vs. \emph{OPC} & vs. \emph{MTL} \\
\midrule
1 & 68.80 (3.98)  & -0.02 (0.07)& -0.07 (0.28)& \RTwo{0.16 (0.24)}& \PST{\ROne{1.47 (1.05)}}& $\PST{+}$& $\PST{+}$& $\PST{+}$\\
2 & 74.83 (3.89)  & \PST{0.64 (0.66)}& \PST{\RTwo{1.12 (0.64)}}& 0.87 (1.53)& \PST{\ROne{2.00 (1.10)}}& $\PST{+}$& $\PST{+}$& $\PST{+}$\\
3 & 78.59 (2.36)  & \PST{0.86 (0.66)}& \PST{1.49 (1.10)}& \PST{\RTwo{1.93 (1.23)}}& \PST{\ROne{2.22 (1.10)}}& $\PST{+}$& $\PST{+}$& $0$\\
4 & 73.92 (2.51)  & -0.17 (0.39)& 0.27 (0.42)& \PST{\ROne{1.82 (1.72)}}& \RTwo{1.38 (2.03)}& $\PST{+}$& $0$& $0$\\
5 & 74.61 (3.37)  & \PST{1.30 (1.35)}& \PST{0.98 (1.35)}& \PST{\RTwo{2.36 (1.46)}}& \PST{\ROne{2.73 (2.01)}}& $\PST{+}$& $\PST{+}$& $0$\\
6 & 63.86 (5.24)  & -0.00 (0.41)& 0.58 (0.91)& \PST{\ROne{1.48 (1.99)}}& \RTwo{0.89 (1.34)}& $0$& $0$& $0$\\
7 & 76.97 (2.21)  & \PST{1.02 (0.70)}& \PST{0.54 (0.34)}& \PST{\ROne{1.18 (0.66)}}& \PST{\RTwo{1.06 (0.79)}}& $0$& $\PST{+}$& $0$\\
8 & 62.97 (3.05)  & -0.00 (0.04)& \RTwo{0.26 (0.45)}& 0.20 (0.49)& \ROne{0.76 (1.11)}& $0$& $0$& $0$\\
9 & 72.52 (2.57)  & \PST{1.05 (0.99)}& \PST{1.06 (0.57)}& \PST{\RTwo{1.53 (1.19)}}& \PST{\ROne{3.08 (1.84)}}& $\PST{+}$& $\PST{+}$& $\PST{+}$\\
10 & 63.62 (2.99)  & 0.09 (0.35)& 0.50 (1.69)& \PST{\RTwo{0.75 (0.86)}}& \PST{\ROne{2.30 (1.47)}}& $\PST{+}$& $0$& $\PST{+}$\\
11 & 59.70 (3.69)  & 0.02 (0.30)& 0.02 (0.29)& \ROne{0.66 (0.98)}& \RTwo{0.54 (1.30)}& $0$& $0$& $0$\\
12 & 71.08 (2.09)  & 0.17 (0.40)& -0.04 (0.74)& \PST{\ROne{0.87 (0.95)}}& \PST{\RTwo{0.87 (0.72)}}& $\PST{+}$& $\PST{+}$& $0$\\
13 & 78.10 (2.31)  & \PST{0.25 (0.31)}& 0.11 (0.41)& \PST{\ROne{1.88 (1.09)}}& \PST{\RTwo{1.31 (1.27)}}& $\PST{+}$& $\PST{+}$& $\NST{-}$\\
14 & 74.13 (1.90)  & \PST{1.08 (0.89)}& \PST{0.48 (0.39)}& \PST{\RTwo{1.59 (1.32)}}& \PST{\ROne{1.85 (1.93)}}& $0$& $\PST{+}$& $0$\\
15 & 72.23 (3.07)  & 0.02 (0.62)& 0.04 (0.30)& \PST{\RTwo{1.25 (0.92)}}& \PST{\ROne{1.84 (1.28)}}& $\PST{+}$& $\PST{+}$& $\PST{+}$\\
16 & 73.32 (1.97)  & \PST{1.02 (1.13)}& \PST{0.67 (0.67)}& \PST{\ROne{1.65 (1.81)}}& \RTwo{1.54 (2.30)}& $0$& $0$& $0$\\
17 & 58.11 (4.61)  & 0.09 (0.15)& 0.18 (0.39)& \PST{\RTwo{0.51 (0.61)}}& \PST{\ROne{1.16 (0.95)}}& $\PST{+}$& $\PST{+}$& $\PST{+}$\\
18 & 57.35 (4.92)  & -0.04 (0.23)& 0.29 (0.65)& \RTwo{0.55 (0.92)}& \ROne{0.71 (1.85)}& $0$& $0$& $0$\\
19 & 76.67 (3.06)  & \PST{1.28 (1.15)}& 0.43 (0.98)& \PST{\RTwo{1.30 (0.94)}}& \PST{\ROne{1.92 (1.47)}}& $0$& $\PST{+}$& $0$\\
20 & 76.31 (2.10)  & 0.28 (0.47)& -0.06 (0.48)& \ROne{0.81 (1.37)}& \RTwo{0.72 (1.53)}& $0$& $0$& $0$\\
21 & 75.45 (3.03)  & \PST{1.21 (0.93)}& \PST{1.35 (1.54)}& \PST{\RTwo{1.88 (1.55)}}& \PST{\ROne{2.28 (1.50)}}& $0$& $0$& $0$\\
22 & 75.28 (4.13)  & \PST{1.01 (0.64)}& 0.49 (0.84)& \PST{\ROne{1.66 (1.08)}}& \PST{\RTwo{1.30 (1.45)}}& $0$& $\PST{+}$& $0$\\
23 & 69.67 (3.00)  & 0.12 (0.43)& 0.02 (0.72)& \PST{\RTwo{1.99 (1.36)}}& \PST{\ROne{2.07 (1.74)}}& $\PST{+}$& $\PST{+}$& $0$\\
24 & 76.24 (2.55)  & \PST{0.87 (0.57)}& \PST{\RTwo{1.17 (0.87)}}& \PST{0.77 (0.74)}& \PST{\ROne{2.16 (1.49)}}& $\PST{+}$& $0$& $\PST{+}$\\
25 & 70.57 (1.93)  & -0.04 (0.08)& 0.43 (0.76)& \RTwo{0.94 (1.60)}& \PST{\ROne{1.21 (1.20)}}& $\PST{+}$& $\PST{+}$& $0$\\
26 & 63.59 (2.97)  & -0.09 (0.19)& 0.07 (0.42)& \RTwo{0.52 (1.15)}& \PST{\ROne{0.82 (0.53)}}& $\PST{+}$& $\PST{+}$& $0$\\
27 & 72.15 (3.19)  & \RTwo{0.39 (0.56)}& 0.19 (0.49)& -0.06 (0.41)& \PST{\ROne{0.41 (0.49)}}& $0$& $0$& $\PST{+}$\\
28 & 64.40 (3.29)  & 0.00 (0.54)& 0.29 (0.92)& \PST{\RTwo{1.21 (0.97)}}& \PST{\ROne{1.27 (1.27)}}& $\PST{+}$& $\PST{+}$& $0$\\
29 & 57.52 (2.78)  & 0.09 (0.24)& \PST{0.76 (0.96)}& \PST{\RTwo{2.20 (2.42)}}& \PST{\ROne{2.50 (2.58)}}& $\PST{+}$& $\PST{+}$& $0$\\
30 & 56.73 (2.86)  & 0.41 (0.72)& \PST{0.78 (0.93)}& \PST{\ROne{1.98 (1.72)}}& \PST{\RTwo{1.89 (2.30)}}& $0$& $0$& $0$\\
31 & 73.27 (2.96)  & \PST{1.13 (1.11)}& \PST{1.14 (0.67)}& \PST{\RTwo{1.91 (0.97)}}& \PST{\ROne{2.00 (1.09)}}& $\PST{+}$& $\PST{+}$& $0$\\
32 & \RTwo{52.82 (3.71)} & -0.12 (0.53)& -0.12 (0.55)& -0.00 (0.47)& \PST{\ROne{0.64 (0.81)}}& $\PST{+}$& $\PST{+}$& $\PST{+}$\\
33 & 69.13 (2.63)  & 0.14 (0.34)& 0.21 (0.54)& \PST{\ROne{0.52 (0.66)}}& \RTwo{0.27 (0.98)}& $0$& $0$& $0$\\
34 & 58.15 (4.74)  & -0.07 (0.48)& 0.06 (0.44)& \PST{\RTwo{0.46 (0.55)}}& \PST{\ROne{0.90 (1.10)}}& $\PST{+}$& $\PST{+}$& $0$\\
35 & 58.61 (3.93)  & \RTwo{0.52 (1.35)}& 0.48 (1.58)& -0.10 (1.16)& \PST{\ROne{1.64 (1.26)}}& $0$& $\PST{+}$& $\PST{+}$\\
36 & 58.91 (3.40)  & 0.35 (0.59)& 0.15 (0.79)& \RTwo{0.78 (1.84)}& \PST{\ROne{1.73 (1.16)}}& $\PST{+}$& $\PST{+}$& $0$\\
37 & 52.60 (4.12)  & 0.14 (0.42)& 0.08 (0.26)& \RTwo{0.27 (0.38)}& \ROne{0.67 (1.01)}& $0$& $0$& $0$\\
38 & 67.73 (3.67)  & \PST{2.08 (2.28)}& \PST{1.64 (0.93)}& \PST{\ROne{3.37 (2.68)}}& \PST{\RTwo{3.34 (2.47)}}& $\PST{+}$& $\PST{+}$& $0$\\
39 & 76.39 (2.22)  & -0.25 (1.44)& 0.29 (0.41)& \PST{\RTwo{0.67 (0.72)}}& \PST{\ROne{1.19 (0.92)}}& $0$& $\PST{+}$& $\PST{+}$\\
40 & 70.58 (3.12)  & -0.04 (0.28)& 0.09 (0.38)& \PST{\RTwo{1.06 (0.64)}}& \PST{\ROne{1.95 (0.80)}}& $\PST{+}$& $\PST{+}$& $\PST{+}$\\
\bottomrule
\end{tabular}%
}
\end{table*}

\begin{table*}[p]
\caption{\label{t:awa2results1}%
    Ranking accuracies of different predictor combination algorithms on the first 40 attributes of \emph{AWA2} dataset. We repeated experiments 10 times with different training, validation, and test 
set splits.  
    For baseline $\mbf^0$ (second column), Kendall's Tau correlations$\times$100 (standard deviations in parentheses) are presented. For the remaining algorithms (third to sixth columns), the accuracy offsets from $\mbf^0$ are presented.
    The best and second best results are highlighted with \ROne{bold} and \RTwo{italic} fonts, respectively. The results of statistical significance test based on a t--test with $\alpha = 0.95$ are highlighted in \PST{green} (significantly positive) and \NST{orange} (significantly negative). The last three columns show the results of statistical significance test of our algorithm with \emph{GL}, \emph{OPC}, and \emph{MTL}, respectively
	($\PST{+}$/$\NST{-}$: significantly positive/negative).
}
\centering
\resizebox{\linewidth}{!}{%
\renewcommand*{\arraystretch}{0.88}
\setlength{\tabcolsep}{3pt}%
\begin{tabular}{crrrrr|ccc}
\toprule
Attr. & Baseline $\mbf^0$ & \emph{GL} & \emph{OPC} & \emph{MTL} & \emph{NPC (ours)}& vs. \emph{GL} & vs. \emph{OPC} & vs. \emph{MTL} \\
\midrule
1 & 77.86 (3.70)  & \RTwo{0.45 (0.92)}& 0.12 (0.26)& 0.15 (0.29)& \PST{\ROne{7.25 (2.73)}}& $\PST{+}$& $\PST{+}$& $\PST{+}$\\
2 & 83.79 (3.18)  & 0.05 (0.11)& -0.17 (0.41)& \RTwo{0.33 (0.60)}& \PST{\ROne{6.22 (2.14)}}& $\PST{+}$& $\PST{+}$& $\PST{+}$\\
3 & 98.55 (0.65)  & \RTwo{0.02 (0.11)}& \ROne{0.04 (0.07)}& -0.01 (0.06)& 0.02 (0.46)& $0$& $0$& $0$\\
4 & 88.21 (3.47)  & 0.03 (0.32)& \RTwo{0.13 (0.29)}& \PST{0.05 (0.05)}& \PST{\ROne{5.22 (2.27)}}& $\PST{+}$& $\PST{+}$& $\PST{+}$\\
5 & 88.53 (1.90)  & \RTwo{0.21 (0.49)}& 0.12 (0.21)& 0.04 (0.29)& \PST{\ROne{3.55 (2.17)}}& $\PST{+}$& $\PST{+}$& $\PST{+}$\\
6 & \RTwo{97.94 (1.07)} & -0.07 (0.17)& -0.02 (0.07)& -0.12 (0.29)& \PST{\ROne{0.69 (0.65)}}& $\PST{+}$& $\PST{+}$& $\PST{+}$\\
7 & \RTwo{99.22 (0.34)} & -0.05 (0.11)& -0.02 (0.06)& -0.00 (0.09)& \PST{\ROne{0.24 (0.23)}}& $\PST{+}$& $\PST{+}$& $\PST{+}$\\
8 & 82.30 (1.69)  & -0.03 (0.12)& 0.12 (0.39)& \RTwo{0.13 (0.21)}& \PST{\ROne{4.32 (1.88)}}& $\PST{+}$& $\PST{+}$& $\PST{+}$\\
9 & 79.33 (4.37)  & -0.01 (0.35)& \RTwo{0.17 (0.42)}& 0.04 (0.14)& \PST{\ROne{7.05 (1.65)}}& $\PST{+}$& $\PST{+}$& $\PST{+}$\\
10 & 98.58 (0.85)  & \RTwo{0.08 (0.31)}& 0.03 (0.13)& 0.01 (0.03)& \ROne{0.26 (0.40)}& $0$& $0$& $0$\\
11 & 97.44 (0.97)  & -0.03 (0.23)& 0.01 (0.10)& \RTwo{0.06 (0.17)}& \PST{\ROne{0.90 (0.35)}}& $\PST{+}$& $\PST{+}$& $\PST{+}$\\
12 & 94.46 (1.91)  & 0.00 (0.25)& 0.04 (0.31)& \PST{\RTwo{0.47 (0.52)}}& \PST{\ROne{1.13 (0.70)}}& $\PST{+}$& $\PST{+}$& $\PST{+}$\\
13 & 93.52 (1.05)  & -0.14 (0.24)& \RTwo{0.08 (0.31)}& 0.05 (0.19)& \PST{\ROne{2.43 (0.68)}}& $\PST{+}$& $\PST{+}$& $\PST{+}$\\
14 & 94.50 (1.64)  & 0.04 (0.15)& 0.33 (0.49)& \PST{\RTwo{0.55 (0.43)}}& \PST{\ROne{2.29 (0.64)}}& $\PST{+}$& $\PST{+}$& $\PST{+}$\\
15 & 95.04 (1.21)  & 0.00 (0.07)& \RTwo{0.25 (0.40)}& 0.13 (0.19)& \PST{\ROne{1.70 (0.68)}}& $\PST{+}$& $\PST{+}$& $\PST{+}$\\
16 & 85.91 (2.85)  & -0.02 (0.03)& 0.05 (0.23)& \PST{\RTwo{1.39 (1.03)}}& \PST{\ROne{5.88 (2.32)}}& $\PST{+}$& $\PST{+}$& $\PST{+}$\\
17 & 87.00 (2.60)  & \PST{0.48 (0.64)}& 0.01 (0.36)& \PST{\RTwo{1.64 (0.76)}}& \PST{\ROne{4.33 (1.84)}}& $\PST{+}$& $\PST{+}$& $\PST{+}$\\
18 & 99.25 (0.55)  & 0.01 (0.03)& \RTwo{0.08 (0.18)}& 0.02 (0.10)& \ROne{0.18 (0.26)}& $0$& $0$& $0$\\
19 & \ROne{99.75 (0.22)} & \RTwo{-0.00 (0.01)}& -0.02 (0.04)& -0.02 (0.05)& -0.05 (0.31)& $0$& $0$& $0$\\
20 & 97.88 (0.99)  & 0.03 (0.15)& 0.07 (0.26)& \RTwo{0.27 (0.40)}& \PST{\ROne{0.79 (0.59)}}& $\PST{+}$& $\PST{+}$& $\PST{+}$\\
21 & 92.36 (2.33)  & 0.08 (0.13)& \RTwo{0.30 (0.62)}& 0.13 (0.37)& \PST{\ROne{2.44 (1.35)}}& $\PST{+}$& $\PST{+}$& $\PST{+}$\\
22 & 96.81 (1.03)  & 0.01 (0.06)& \RTwo{0.20 (0.35)}& \PST{0.18 (0.18)}& \PST{\ROne{1.40 (0.77)}}& $\PST{+}$& $\PST{+}$& $\PST{+}$\\
23 & 91.60 (2.67)  & 0.11 (0.36)& \RTwo{0.15 (0.23)}& 0.07 (0.14)& \PST{\ROne{3.57 (0.88)}}& $\PST{+}$& $\PST{+}$& $\PST{+}$\\
24 & 95.46 (1.41)  & 0.02 (0.30)& \RTwo{0.06 (0.23)}& 0.01 (0.14)& \PST{\ROne{1.43 (1.24)}}& $\PST{+}$& $\PST{+}$& $\PST{+}$\\
25 & 89.94 (2.52)  & -0.02 (0.08)& 0.07 (0.43)& \RTwo{0.14 (0.45)}& \PST{\ROne{3.55 (1.68)}}& $\PST{+}$& $\PST{+}$& $\PST{+}$\\
26 & 84.78 (3.62)  & 0.03 (0.28)& \RTwo{0.61 (1.16)}& 0.23 (0.34)& \PST{\ROne{3.77 (1.15)}}& $\PST{+}$& $\PST{+}$& $\PST{+}$\\
27 & 90.67 (2.26)  & -0.27 (1.30)& \PST{0.92 (1.09)}& \PST{\RTwo{1.09 (1.05)}}& \PST{\ROne{3.91 (1.80)}}& $\PST{+}$& $\PST{+}$& $\PST{+}$\\
28 & 90.67 (1.67)  & 0.00 (0.03)& \RTwo{0.05 (0.23)}& -0.00 (0.10)& \PST{\ROne{4.51 (1.59)}}& $\PST{+}$& $\PST{+}$& $\PST{+}$\\
29 & 95.73 (1.49)  & \ROne{0.16 (0.30)}& -0.04 (0.16)& -0.11 (0.34)& \RTwo{0.15 (0.80)}& $0$& $0$& $0$\\
30 & 94.97 (1.58)  & 0.04 (0.11)& \RTwo{0.12 (0.42)}& 0.06 (0.17)& \PST{\ROne{1.21 (1.10)}}& $\PST{+}$& $\PST{+}$& $\PST{+}$\\
31 & 94.39 (2.23)  & \PST{\RTwo{0.41 (0.45)}}& \PST{0.40 (0.27)}& \PST{0.41 (0.48)}& \PST{\ROne{2.52 (0.99)}}& $\PST{+}$& $\PST{+}$& $\PST{+}$\\
32 & 96.92 (1.17)  & -0.02 (0.13)& \RTwo{0.14 (0.33)}& 0.04 (0.11)& \PST{\ROne{0.86 (0.86)}}& $\PST{+}$& $0$& $\PST{+}$\\
33 & 85.97 (4.63)  & 0.08 (0.21)& \RTwo{0.16 (0.53)}& 0.02 (0.20)& \PST{\ROne{6.68 (3.30)}}& $\PST{+}$& $\PST{+}$& $\PST{+}$\\
34 & 98.58 (0.83)  & 0.00 (0.01)& \RTwo{0.07 (0.14)}& -0.07 (0.12)& \ROne{0.25 (0.60)}& $0$& $0$& $0$\\
35 & 98.69 (0.47)  & 0.03 (0.07)& \PST{\RTwo{0.30 (0.34)}}& \PST{0.26 (0.28)}& \PST{\ROne{0.45 (0.45)}}& $\PST{+}$& $0$& $0$\\
36 & \RTwo{97.71 (0.82)} & -0.01 (0.14)& -0.02 (0.15)& -0.07 (0.13)& \PST{\ROne{0.44 (0.27)}}& $\PST{+}$& $\PST{+}$& $\PST{+}$\\
37 & 97.19 (1.48)  & -0.04 (0.12)& 0.22 (0.59)& \PST{\RTwo{0.34 (0.42)}}& \PST{\ROne{1.17 (1.18)}}& $\PST{+}$& $\PST{+}$& $\PST{+}$\\
38 & 95.21 (0.97)  & \RTwo{0.05 (0.10)}& -0.00 (0.16)& 0.04 (0.17)& \PST{\ROne{1.95 (0.75)}}& $\PST{+}$& $\PST{+}$& $\PST{+}$\\
39 & 89.66 (1.83)  & -0.02 (0.05)& 0.13 (0.75)& \PST{\RTwo{1.17 (0.94)}}& \PST{\ROne{4.01 (1.53)}}& $\PST{+}$& $\PST{+}$& $\PST{+}$\\
40 & 95.33 (1.58)  & -0.04 (0.17)& \PST{0.20 (0.26)}& \PST{\RTwo{0.64 (0.45)}}& \PST{\ROne{2.60 (1.27)}}& $\PST{+}$& $\PST{+}$& $\PST{+}$\\
\bottomrule
\end{tabular}%
}
\end{table*}

\begin{table*}[p]
\caption{\label{t:awa2results2}%
    Ranking accuracies of different predictor combination algorithms on the last 40 attributes of \emph{AWA2} dataset. For each dataset, we repeated experiments 10 times with different training, validation, and test set splits.  
    For baseline $\mbf^0$ (second column), Kendall's Tau correlations$\times$100 (standard deviations in parentheses) are presented. For the remaining algorithms (third to sixth columns), the accuracy offsets from $\mbf^0$ are presented.
    The best and second best results are highlighted with \ROne{bold} and \RTwo{italic} fonts, respectively. The results of statistical significance test based on a t--test with $\alpha = 0.95$ are highlighted in \PST{green} (significantly positive) and \NST{orange} (significantly negative). The last three columns show the results of statistical significance test of our algorithm with \emph{GL}, \emph{OPC}, and \emph{MTL}, respectively
	($\PST{+}$/$\NST{-}$: significantly positive/negative).
}
\centering
\resizebox{\linewidth}{!}{%
\renewcommand*{\arraystretch}{0.88}
\setlength{\tabcolsep}{3pt}%
\begin{tabular}{crrrrr|ccc}
\toprule
Attr. & Baseline $\mbf^0$ & \emph{GL} & \emph{OPC} & \emph{MTL} & \emph{NPC (ours)}& vs. \emph{GL} & vs. \emph{OPC} & vs. \emph{MTL} \\
\midrule
41 & 95.21 (1.26)  & -0.06 (0.16)& 0.01 (0.13)& \RTwo{0.07 (0.19)}& \PST{\ROne{1.03 (0.60)}}& $\PST{+}$& $\PST{+}$& $\PST{+}$\\
42 & 84.32 (3.88)  & \RTwo{0.21 (0.49)}& -0.06 (0.17)& -0.07 (0.27)& \PST{\ROne{2.86 (1.72)}}& $\PST{+}$& $\PST{+}$& $\PST{+}$\\
43 & 95.78 (1.49)  & 0.06 (0.12)& -0.10 (0.19)& \RTwo{0.13 (0.20)}& \PST{\ROne{2.57 (1.12)}}& $\PST{+}$& $\PST{+}$& $\PST{+}$\\
44 & \RTwo{98.22 (0.78)} & -0.05 (0.25)& -0.00 (0.04)& -0.00 (0.17)& \PST{\ROne{0.62 (0.50)}}& $\PST{+}$& $\PST{+}$& $\PST{+}$\\
45 & 88.71 (2.33)  & 0.07 (0.20)& 0.42 (0.65)& \PST{\RTwo{0.65 (0.83)}}& \PST{\ROne{2.92 (2.09)}}& $\PST{+}$& $\PST{+}$& $\PST{+}$\\
46 & 83.63 (2.01)  & 0.01 (0.04)& 0.43 (0.74)& \PST{\RTwo{0.77 (0.41)}}& \PST{\ROne{8.15 (1.08)}}& $\PST{+}$& $\PST{+}$& $\PST{+}$\\
47 & 89.57 (2.34)  & \RTwo{0.39 (0.59)}& 0.16 (0.33)& 0.13 (0.34)& \PST{\ROne{3.66 (2.15)}}& $\PST{+}$& $\PST{+}$& $\PST{+}$\\
48 & 91.92 (2.16)  & \PST{0.18 (0.24)}& 0.51 (0.83)& \PST{\RTwo{0.51 (0.43)}}& \PST{\ROne{2.70 (1.81)}}& $\PST{+}$& $\PST{+}$& $\PST{+}$\\
49 & 89.56 (3.03)  & 0.05 (0.27)& -0.03 (0.25)& \PST{\RTwo{0.34 (0.40)}}& \PST{\ROne{3.65 (1.79)}}& $\PST{+}$& $\PST{+}$& $\PST{+}$\\
50 & 95.20 (1.24)  & -0.01 (0.05)& -0.06 (0.25)& \RTwo{0.11 (0.21)}& \PST{\ROne{2.23 (0.82)}}& $\PST{+}$& $\PST{+}$& $\PST{+}$\\
51 & 90.78 (2.72)  & -0.34 (1.72)& \RTwo{1.01 (1.61)}& \PST{0.91 (0.81)}& \PST{\ROne{3.16 (1.46)}}& $\PST{+}$& $\PST{+}$& $\PST{+}$\\
52 & 94.00 (1.48)  & -0.14 (0.19)& 0.08 (0.70)& \PST{\RTwo{0.28 (0.29)}}& \PST{\ROne{3.11 (0.97)}}& $\PST{+}$& $\PST{+}$& $\PST{+}$\\
53 & 94.51 (1.71)  & \RTwo{0.10 (0.17)}& -0.00 (0.17)& -0.02 (0.18)& \PST{\ROne{1.25 (0.78)}}& $\PST{+}$& $\PST{+}$& $\PST{+}$\\
54 & 91.17 (1.73)  & 0.15 (0.39)& \RTwo{0.22 (0.51)}& 0.16 (0.47)& \PST{\ROne{4.19 (1.20)}}& $\PST{+}$& $\PST{+}$& $\PST{+}$\\
55 & 93.07 (1.87)  & 0.07 (0.29)& -0.00 (0.38)& \PST{\RTwo{0.30 (0.33)}}& \PST{\ROne{2.95 (0.86)}}& $\PST{+}$& $\PST{+}$& $\PST{+}$\\
56 & 88.30 (3.12)  & 0.29 (0.51)& 0.57 (1.21)& \PST{\RTwo{1.26 (1.36)}}& \PST{\ROne{3.32 (2.47)}}& $\PST{+}$& $\PST{+}$& $\PST{+}$\\
57 & 92.34 (2.15)  & \RTwo{0.34 (0.61)}& 0.03 (0.30)& 0.05 (0.18)& \PST{\ROne{2.59 (0.92)}}& $\PST{+}$& $\PST{+}$& $\PST{+}$\\
58 & 96.10 (1.78)  & \PST{0.16 (0.18)}& 0.11 (0.17)& \RTwo{0.43 (0.75)}& \PST{\ROne{1.13 (1.02)}}& $\PST{+}$& $\PST{+}$& $\PST{+}$\\
59 & 92.12 (3.24)  & \RTwo{0.30 (0.42)}& 0.16 (0.25)& \PST{0.23 (0.23)}& \PST{\ROne{3.07 (1.47)}}& $\PST{+}$& $\PST{+}$& $\PST{+}$\\
60 & 89.37 (1.95)  & 0.12 (0.25)& \RTwo{0.44 (0.75)}& 0.14 (0.23)& \PST{\ROne{2.37 (1.45)}}& $\PST{+}$& $\PST{+}$& $\PST{+}$\\
61 & 92.43 (1.97)  & -0.10 (0.22)& 0.04 (0.27)& \RTwo{0.17 (0.28)}& \PST{\ROne{2.37 (0.88)}}& $\PST{+}$& $\PST{+}$& $\PST{+}$\\
62 & 98.16 (1.08)  & 0.08 (0.27)& 0.07 (0.24)& \PST{\RTwo{0.18 (0.23)}}& \PST{\ROne{0.39 (0.53)}}& $0$& $0$& $0$\\
63 & 90.95 (3.52)  & -0.04 (0.10)& \RTwo{0.33 (0.51)}& 0.15 (0.49)& \PST{\ROne{3.91 (2.55)}}& $\PST{+}$& $\PST{+}$& $\PST{+}$\\
64 & 92.31 (2.46)  & -0.06 (0.13)& \PST{0.19 (0.24)}& \PST{\RTwo{0.39 (0.52)}}& \PST{\ROne{2.80 (1.80)}}& $\PST{+}$& $\PST{+}$& $\PST{+}$\\
65 & 90.41 (3.04)  & 0.14 (0.30)& -0.01 (0.29)& \PST{\RTwo{0.57 (0.49)}}& \PST{\ROne{3.53 (1.25)}}& $\PST{+}$& $\PST{+}$& $\PST{+}$\\
66 & 89.81 (2.71)  & \PST{\RTwo{0.80 (0.89)}}& 0.58 (1.30)& 0.00 (0.27)& \PST{\ROne{3.12 (1.38)}}& $\PST{+}$& $\PST{+}$& $\PST{+}$\\
67 & 94.72 (2.26)  & -0.02 (0.13)& \RTwo{0.11 (0.77)}& 0.07 (0.15)& \PST{\ROne{2.38 (0.83)}}& $\PST{+}$& $\PST{+}$& $\PST{+}$\\
68 & 85.60 (2.43)  & 0.18 (0.83)& \RTwo{0.47 (1.06)}& \PST{0.11 (0.14)}& \PST{\ROne{6.41 (1.36)}}& $\PST{+}$& $\PST{+}$& $\PST{+}$\\
69 & 98.79 (0.41)  & 0.02 (0.05)& 0.14 (0.27)& \RTwo{0.14 (0.22)}& \PST{\ROne{0.50 (0.41)}}& $\PST{+}$& $\PST{+}$& $\PST{+}$\\
70 & 97.09 (1.19)  & -0.04 (0.12)& \PST{0.14 (0.18)}& \PST{\RTwo{0.27 (0.35)}}& \PST{\ROne{0.93 (0.83)}}& $\PST{+}$& $\PST{+}$& $\PST{+}$\\
71 & 99.02 (0.47)  & 0.01 (0.02)& \PST{\RTwo{0.26 (0.20)}}& \PST{0.16 (0.11)}& \PST{\ROne{0.49 (0.32)}}& $\PST{+}$& $\PST{+}$& $\PST{+}$\\
72 & 96.69 (0.57)  & -0.11 (0.29)& \RTwo{0.23 (0.44)}& \PST{0.18 (0.19)}& \PST{\ROne{1.79 (0.72)}}& $\PST{+}$& $\PST{+}$& $\PST{+}$\\
73 & 94.18 (1.34)  & \RTwo{0.09 (0.28)}& -0.00 (0.06)& 0.00 (0.13)& \PST{\ROne{2.48 (0.98)}}& $\PST{+}$& $\PST{+}$& $\PST{+}$\\
74 & 83.62 (3.96)  & -0.02 (0.41)& 0.16 (0.84)& \PST{\RTwo{0.56 (0.62)}}& \PST{\ROne{5.06 (3.19)}}& $\PST{+}$& $\PST{+}$& $\PST{+}$\\
75 & 82.00 (3.99)  & -0.06 (0.11)& \PST{0.54 (0.64)}& \PST{\RTwo{1.43 (1.62)}}& \PST{\ROne{6.32 (2.62)}}& $\PST{+}$& $\PST{+}$& $\PST{+}$\\
76 & 83.13 (3.21)  & 0.08 (0.45)& 0.23 (0.64)& \RTwo{0.29 (0.65)}& \PST{\ROne{6.38 (2.76)}}& $\PST{+}$& $\PST{+}$& $\PST{+}$\\
77 & 85.16 (1.97)  & 0.09 (0.29)& 0.08 (0.40)& \PST{\RTwo{0.76 (0.67)}}& \PST{\ROne{5.58 (2.22)}}& $\PST{+}$& $\PST{+}$& $\PST{+}$\\
78 & 85.83 (2.74)  & 0.14 (0.48)& -0.11 (0.50)& \RTwo{0.43 (0.81)}& \PST{\ROne{4.48 (2.19)}}& $\PST{+}$& $\PST{+}$& $\PST{+}$\\
79 & 87.37 (2.77)  & 0.08 (0.34)& 0.05 (0.29)& \PST{\RTwo{0.50 (0.30)}}& \PST{\ROne{5.25 (1.86)}}& $\PST{+}$& $\PST{+}$& $\PST{+}$\\
80 & 85.30 (2.93)  & 0.04 (0.21)& 0.18 (0.53)& \PST{\RTwo{0.44 (0.30)}}& \PST{\ROne{6.05 (2.39)}}& $\PST{+}$& $\PST{+}$& $\PST{+}$\\
\bottomrule
\end{tabular}%
}
\end{table*}



\end{document}